\documentclass[letterpaper, preprint,11pt]{AAS}	

\usepackage{bm}
\usepackage{amsmath}
\usepackage{graphicx}
\usepackage{float}
\usepackage[colorlinks=true, pdfstartview=FitV, linkcolor=black, citecolor= black, urlcolor= black]{hyperref}
\usepackage{overcite}
\usepackage{stackengine}
\usepackage{subfigure}
\usepackage{footnpag}			      	
\graphicspath{{Figures/}}

\PaperNumber{22-105}

\begin{document}

\title{HAZARD DETECTION AND AVOIDANCE FOR THE NOVA-C LANDER}

\author{Joel Getchius\thanks{Senior Navigation Systems Engineer, Omitron Inc., 7051 Muirkirk Meadows Drive, Suite A, Beltsville, MD 20705.},  
Devin Renshaw\thanks{Software Developer, Intuitive Machines, 3700 Bay Area Blvd, Houston, TX, 77058.},
Daniel Posada\thanks{PhD Candidate, Department of Aerospace Engineering, Embry-Riddle Aeronautical University, 1 Aerospace Boulevard, Daytona Beach, FL, 32114.},
Troy Henderson\thanks{Associate Professor of Aerospace Engineering, Department of Aerospace Engineering, Embry-Riddle Aeronautical University, 1 Aerospace Boulevard, Daytona Beach, FL, 32114.},
Lillian Hong\thanks{Graphics Programmer, Intuitive Machines, 3700 Bay Area Blvd, Houston, TX, 77058.},
Shen Ge\thanks{Integration Engineer, Iron Ring, 2723 Austin St, Houston, TX, 77004.},
\ and Giovanni Molina\thanks{Software Developer, Intuitive Machines, 3700 Bay Area Blvd, Houston, TX, 77058.}
}

\maketitle{} 		

\begin{abstract}
In early 2022, Intuitive Machines’ NOVA-C Lander will touch down on the lunar surface becoming the first commercial endeavor to visit a celestial body.  NOVA-C will deliver six payloads to the lunar surface with various scientific and engineering objectives, ushering in a new era of commercial space exploration and utilization.  However, to safely accomplish the mission, the NOVA-C lander must ensure its landing site is free of hazards larger than 30 cm and the slope of local terrain at touchdown is less than 10 degrees off vertical. To accomplish this, NOVA-C utilizes Intuitive Machines’ precision navigation system, coupled with machine vision algorithms for scene reduction and landing site characterization. A unique aspect to the NOVA-C approach is the real-time nature of the hazard detection and avoidance algorithms--which are performed 400 meters above and down range of the intended landing site and completed within 15 seconds. In this paper, we review the theoretical foundations for the hazard detection and avoidance algorithms, describe the practical challenges of implementation on the NOVA-C flight computer, and present test and analysis results.
\end{abstract}

\section{Introduction}
Scheduled for launch in early 2022, the Intuitive Machines' (IM) NOVA-C lander will deliver six payloads to the lunar surface.  After launch, the NOVA-C will embark on a multi-day journey until insertion into lunar orbit via the lunar orbit insertion burn.  For the next 24 hours the vehicle will remain in a low lunar orbit until the de-orbit initialization burn.  Approximately one hour later, NOVA-C will begin its powered descent initialization which further lowers the orbit and steers the vehicle to a vertical descent to the intended landing site.

To ensure a safe landing, NOVA-C requires a landing site no smaller than 10 m by 10 m, free of hazards larger than 30 cm, and with a slope less than 10 degrees off horizontal.  Unfortunately, a priori maps of the region encompassing the intended landing site typically have, at best, 1 meter per pixel resolution\cite{baker2015}.  Clearly, this is insufficient to ensure the landing site requirements and thus real-time hazard detection and avoidance (HDA) algorithms are necessary.  

\begin{figure}[htb]
	\centering\includegraphics[width=3.5in]{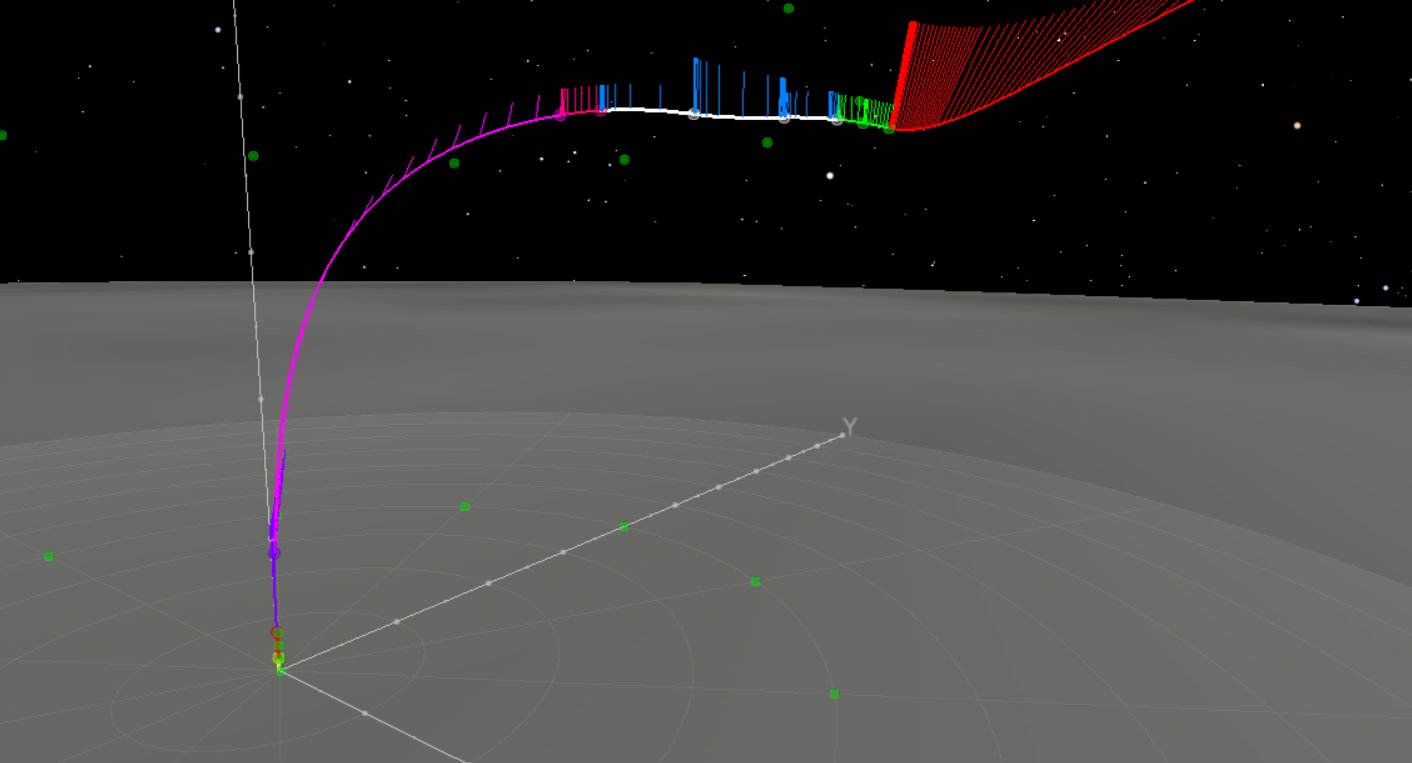}
	\caption{Nominal landing trajectory with respect to the ILS.  During the green highlight, HDA performs image collection.  White is contingency image collection and time for HDA to complete computations.}
	\label{fig:hda_traj}
\end{figure}

Figure~\ref{fig:hda_traj} illustrates the operational trajectory for HDA, which  begins at 400 meters altitude and 400 meters downrange from the intended landing site (ILS).  The concept of operations allot approximately 5 seconds for image collection (green) and an additional 10 seconds for computational time (white).  A significant challenge is completing the HDA algorithms in the total allotted 15 seconds all on a processor with 1 GHz capability.

Architecturally, the NOVA-C lander has two computing platforms:

\begin{enumerate}
  \item A central processing unit (CPU) for the guidance, navigation, and control (GNC) algorithms.
  \item A vision processing unit (VPU) for the image processing algorithms such as HDA.
\end{enumerate}

For the first NOVA-C landing, an optical camera is the primary sensor. Other landing programs have investigated the use of lidar-based surface slope detection\cite{esa_landing}. Furthermore, the NOVA-C navigation system consists of two, independent inertial measurement units (IMUs) and a laser range finder that is co-boresighted with an optical camera.  Two inertial star trackers provide precision measurements of attitude.  A Kalman filter fuses all measurement types (laser range finder, visual odometry, crater tracking, IMU, and star tracker) to provide the rest of the spacecraft systems estimates of position, velocity and attitude.  The HDA system, in particular, depends on time stamped imagery with corresponding navigation data (for relative motion), and laser range finder measurements for scale in its algorithms. The time tag of imagery is synchronized on capture with the flight computer and navigation filter and is stored as part of the image packet. It is worth noting that while the NOVA-C HDA concept of operations makes use of the navigation data for relative motion and attitude information, during HDA only the IMU is in use by the navigation system.  This propagation only mode, protects the HDA algorithms from the natural discontinuities that exist between sensor fusion and filtering algorithms utilized in the navigation system.

The computationally complex, and specific, HDA requirements necessitate a series of test campaigns to ensure correct implementation and give a high degree of confidence in the HDA system to ensure a successful lunar landing.  Testing includes software in the loop (SIL), processor in the loop (PIL), and several instantiations of field testing.  

Field testing in particular presented a unique suite of challenges that had to be addressed in order for the tests to increase the technical readiness level for a lunar mission.  First and foremost, is the need for a sensor suite in as much of a flight like configuration as possible.  The NOVA-C team invested significant resources in the construction of ``Navpod": a cubical structure housing engineering units of the flight computational platforms, the laser range finder, IMUs, and an optical camera.

In this paper, we discuss the development of the HDA algorithms and describe completed testing campaigns.  Additionally we preview upcoming test campaigns and describe the future work needed to complete the NOVA-C HDA implementation, verification, and test.

\section{HDA Algorithm Design}

Leveraging off of previous HDA development efforts\cite{epp2007autonomous,crane2014vision} and a recognition that any HDA implementation is unique to the landing requirements, available sensors, and computational requirements\cite{villalpando2010investigation} of a particular mission, researchers at Embry-Riddle Aeronautical University developed a suite of algorithms for the NOVA-C HDA system. The only current active implemented HDA development was used for Mars 2020 Perseverance Rover \cite{aaron2022camera}, where they leveraged their Terrain Relative Navigation (TRN) technology and combined with a landing camera that required to resize the imagery to be able to processed in time in par with a loaded on board map that required some time for planning and development\cite{cheng2021making}.

Based initially on the HDA algorithm developed by Posada\cite{posada2020a}, the algorithm uses a passive camera sensor, an IMU, and an active laser range finder. This method is significantly lighter computationally, but meets requirements for NOVA-C landing. The algorithm initially incorporated two layers of segmentation (superpixels and quadtree) for hazard detection.  Computational limitations on spaceflight computer systems led to an alteration where the superpixel segmentation was abandoned.  Therefore, the segmentation was reduced to use only the quadtree approach to generate a occupancy grid based on illuminated and shadowed regions.

The modified HDA algorithm consists of two major parts:

\begin{itemize}
  \item Quadtree decomposition of imagery and identification of potential landing sites.
  \item Structure from motion generates point clouds of landing sites of interest to evaluate for slope and roughness.
\end{itemize}

\begin{figure}[ht]
    \centering
    \begin{minipage}[t]{0.3\linewidth}
        \centering
        \includegraphics[width=\linewidth]{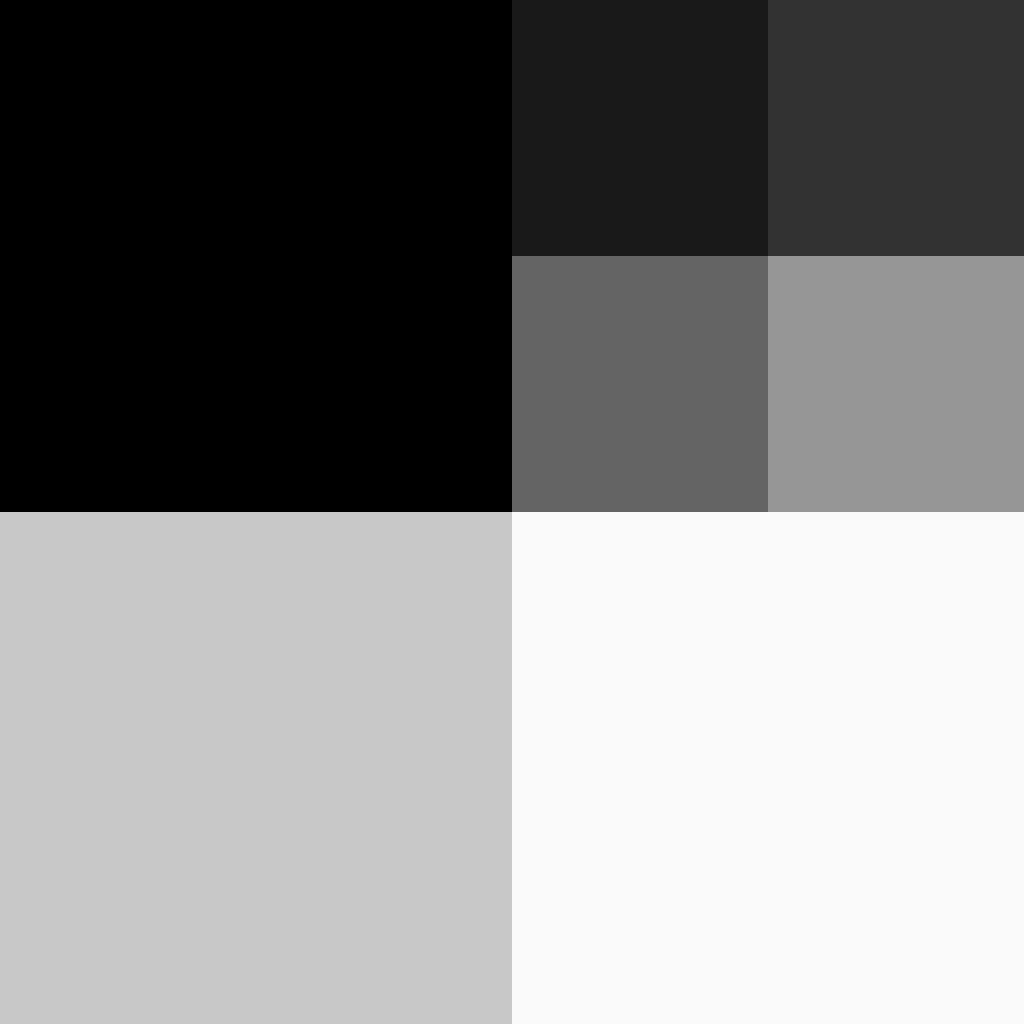}
    \end{minipage}
    \begin{minipage}[t]{0.3\textwidth}
        \centering
        \includegraphics[width=\linewidth]{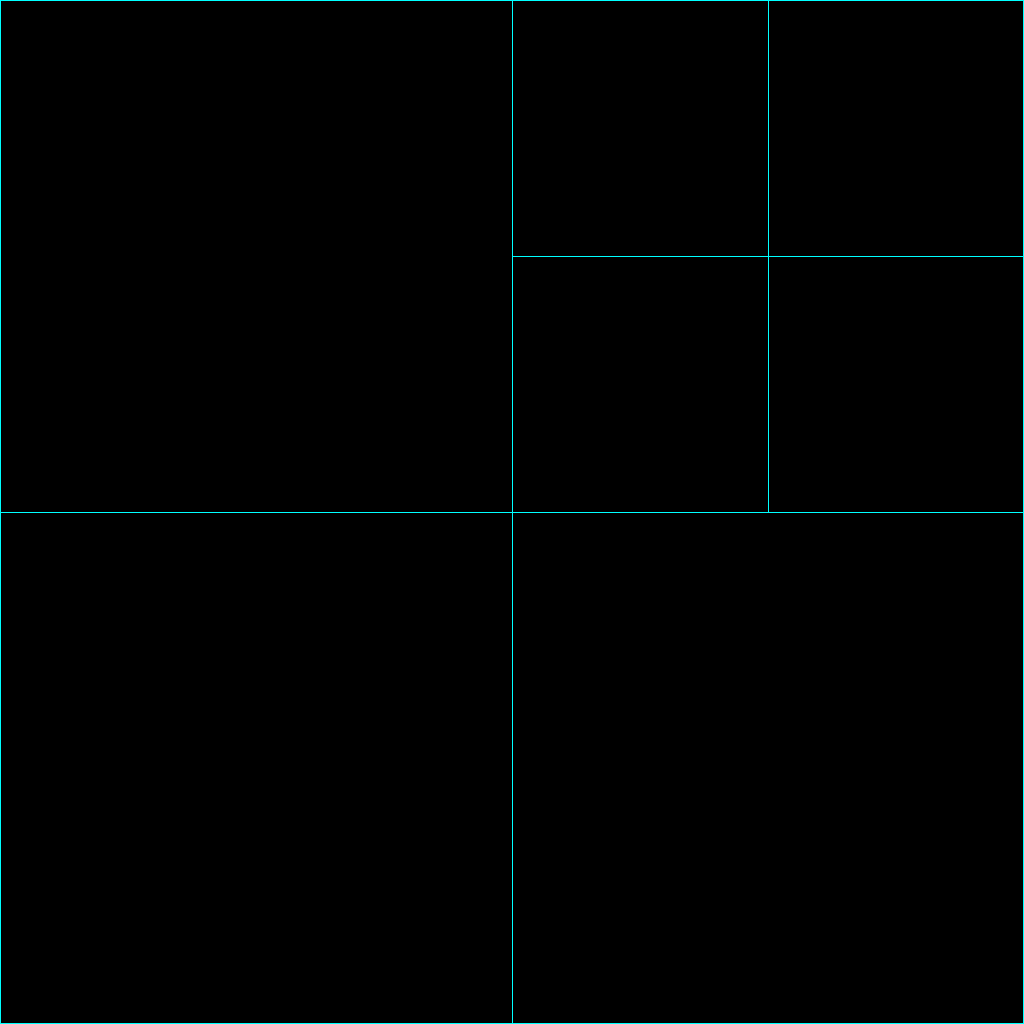}
    \end{minipage}
    \begin{minipage}[t]{0.3\textwidth}
        \centering
        \includegraphics[width=\linewidth]{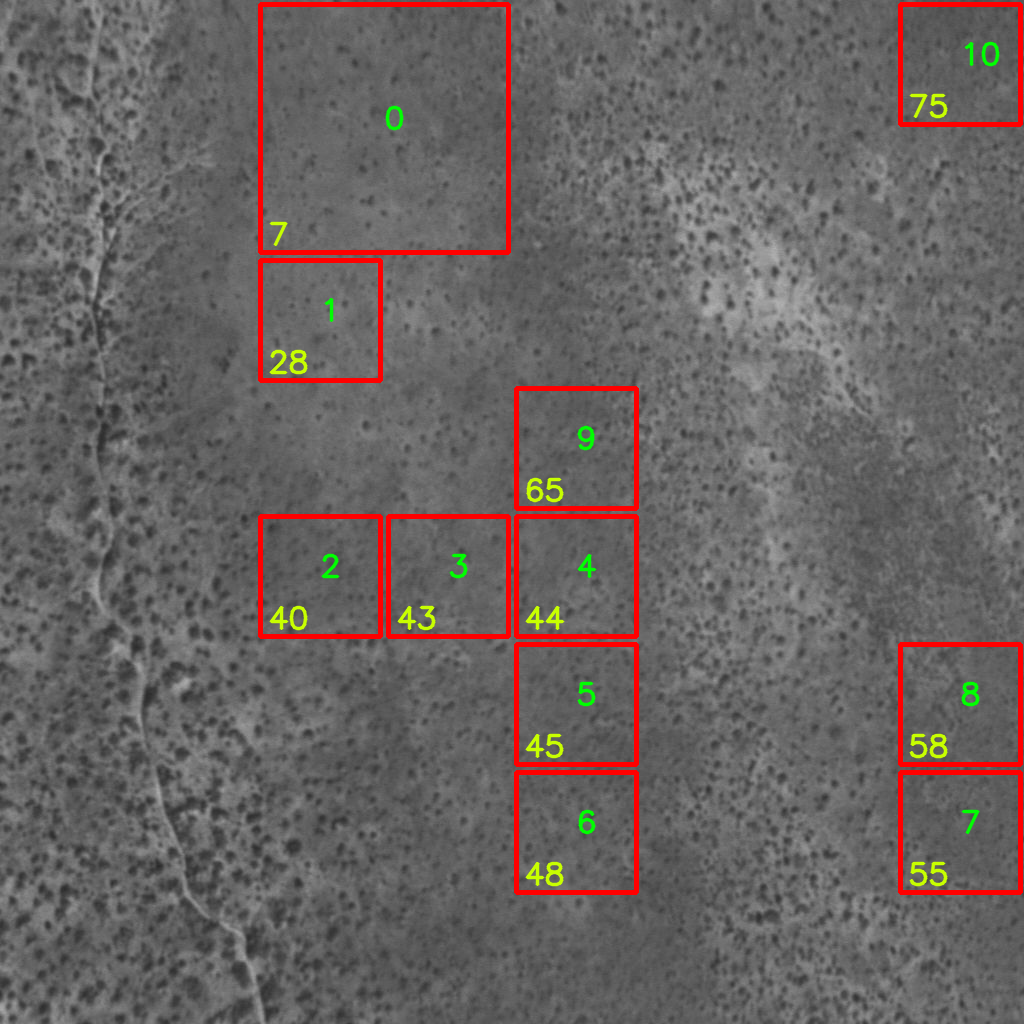}
    \end{minipage}
    \caption{Left: Sample with different areas and intensities\cite{posada2020a}. Center: Quadtree decomposition\cite{posada2020a}. Right: Example of Quadtree candidates after applied filters.}
    \label{fig:opencv:quadtree}
\end{figure}

The quadtree decomposition reduces the image to candidate landing sites by sub-dividing regions based on the standard deviation of brightness in the region. An illuminated area will be imaged as a bright region while a region in shadow will show as dark--indicating the presence of hazards casting the shadow (whether rocks or craters). The minimum size of allowed decomposition is just under the required landing are (10x10 m$^2$). An example is seen in Figure \ref{fig:opencv:quadtree}. 

The quadtree algorithm decomposes these regions based on the user-defined criteria, specific in this case to landing criteria. To remove hazards, dark areas and areas with large variability in brightness are discarded by analyzing the mean, the standard deviation, and the size of the area. This new map becomes an occupancy grid allowing the choice of a desired region to be further analyzed (right panel of Figure \ref{fig:opencv:quadtree}). The size of the regions is determined by the pixel size and use of a co-boresighted laser range finder to remove range ambiguity. When a candidate region has low lighting variability (image intensity) it is an indicator that the region is low in significant hazards. The regions identified that meet the criteria are labeled as regions of interest (ROIs).

Once the initial filter is performed, reducing the search area to those free of significant hazards in areas that meet the minimum landing area, the remaining candidate landing sites can be further analyzed to ensure they meet slope and roughness requirements. This is accomplished by reconstructing a sparse 3-D topography map based on a point cloud. This process is denoted as Structure From Motion (SfM). 

A point cloud is generated by triangulating and matching features. These features are obtained by finding unique points in the image using a feature detection algorithm. There are multiple known methods such as SIFT, SURF, Fast, but for this approach ORB\cite{karami2017image} was used. The choice of ORB was based on its use in other Nova-C navigation routines, feature detection and description resilience, and computation speed. Once these features are identified in each region of interest, they can be triangulated using epipolar geometry \cite{Hartley2004,epipolar} and are mapped from the pixel space into the 3-D world space.

\begin{figure}[h!]
    \centering
    \begin{minipage}[t]{0.36\linewidth}
        \centering
        \includegraphics[width=\linewidth]{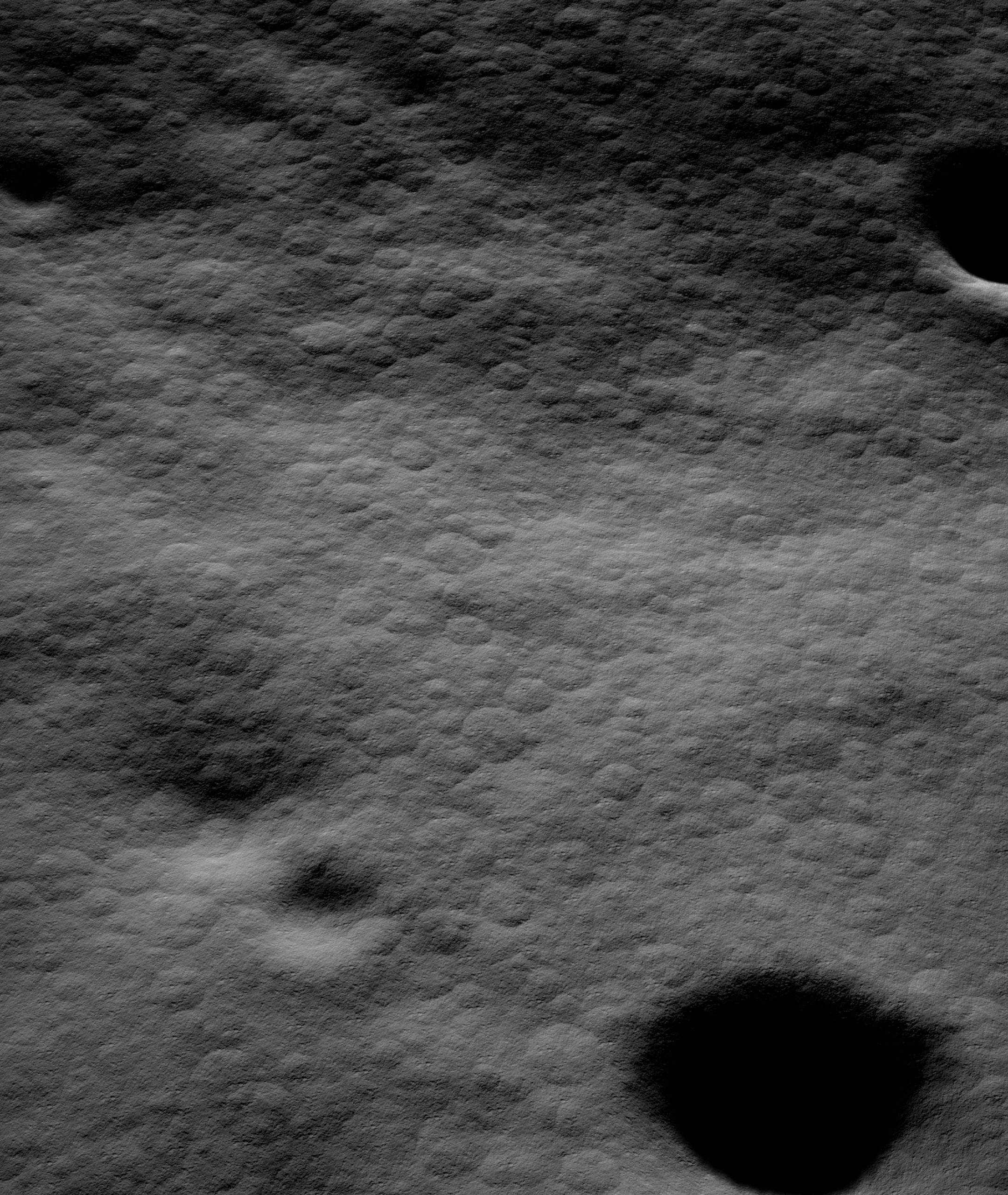}
    \end{minipage}
    \begin{minipage}[t]{0.55\textwidth}
        \centering
        \includegraphics[width=\linewidth]{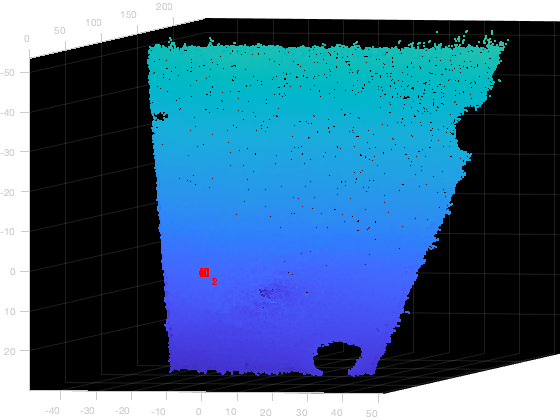}
    \end{minipage}
    \caption{Left: Original lunar surface. Right: Point cloud reconstructed using SfM\cite{posada2020a}.}
    \label{fig:sfm:example}
\end{figure}

Figure \ref{fig:sfm:example} shows an example of the point cloud reconstruction on right side and the original lunar surface. In this particular example, the reconstruction (right panel) is trapezoidal due to the angle which the synthetic picture was taken with respect to the normal of the surface and the camera geometry.

One of the main challenges in the design was to adapt the algorithm to the trajectory of NOVA-C. Therefore, the terminal trajectory was optimized for HDA operations in order to maximize observability in the system (recall that only two images are used along with the IMU and laser range finder). Of particular concern was ensuring the geometric observability in order to adequately reconstruct the 3-D point cloud necessary for the slope and roughness computations. 
The initial trajectory was a glide slope with the camera consistently boresighted on the ILS.  However, geometric observability analysis and Monte Carlo simulations using synthetic features on a controlled terrain to perform structure for motion and therefore point cloud reconstruction revealed its weakness in observing the point cloud.

\begin{figure}[ht]
    \centering\includegraphics[width=2.5in]{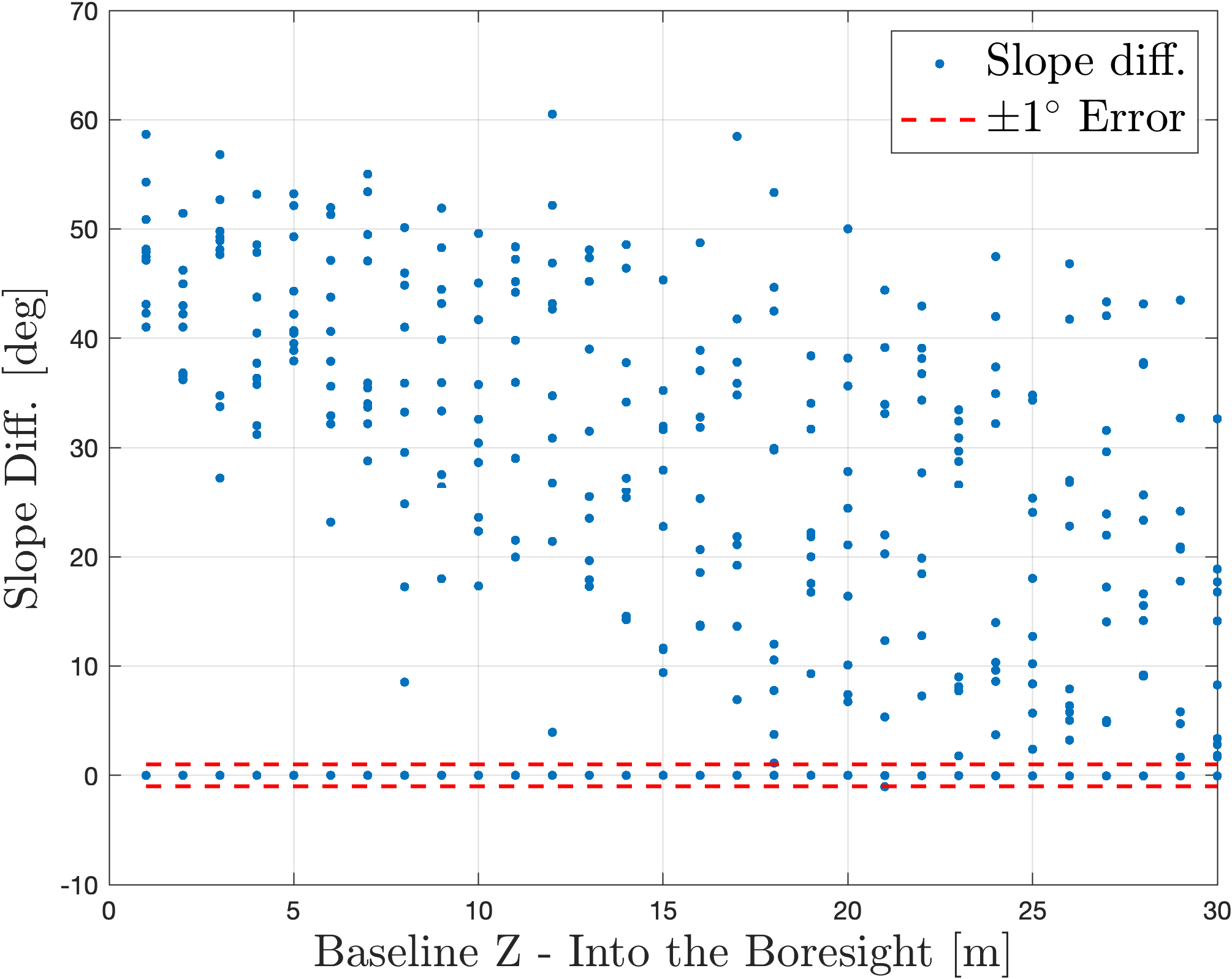}
	\centering\includegraphics[width=2.5in]{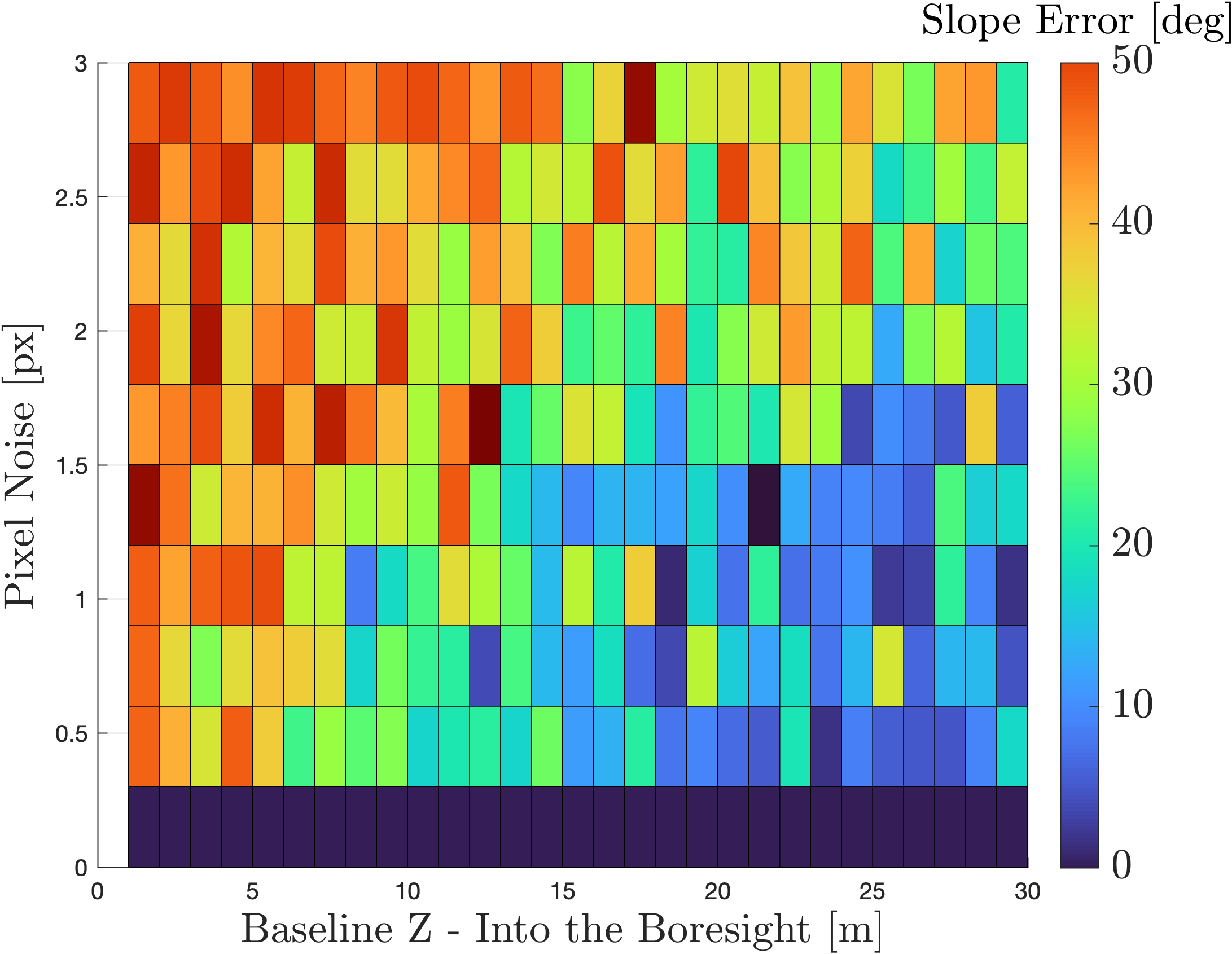}
	\caption{Slope error from the Monte Carlo simulation with synthetic features for relative motion along the camera boresight  Left: Slope error versus change in baseline. Right: Pixel Noise and Slope error versus change in baseline.}
	\label{fig:pc_bad}
\end{figure}

\begin{figure}[H]
	\centering\includegraphics[width=0.8\textwidth]{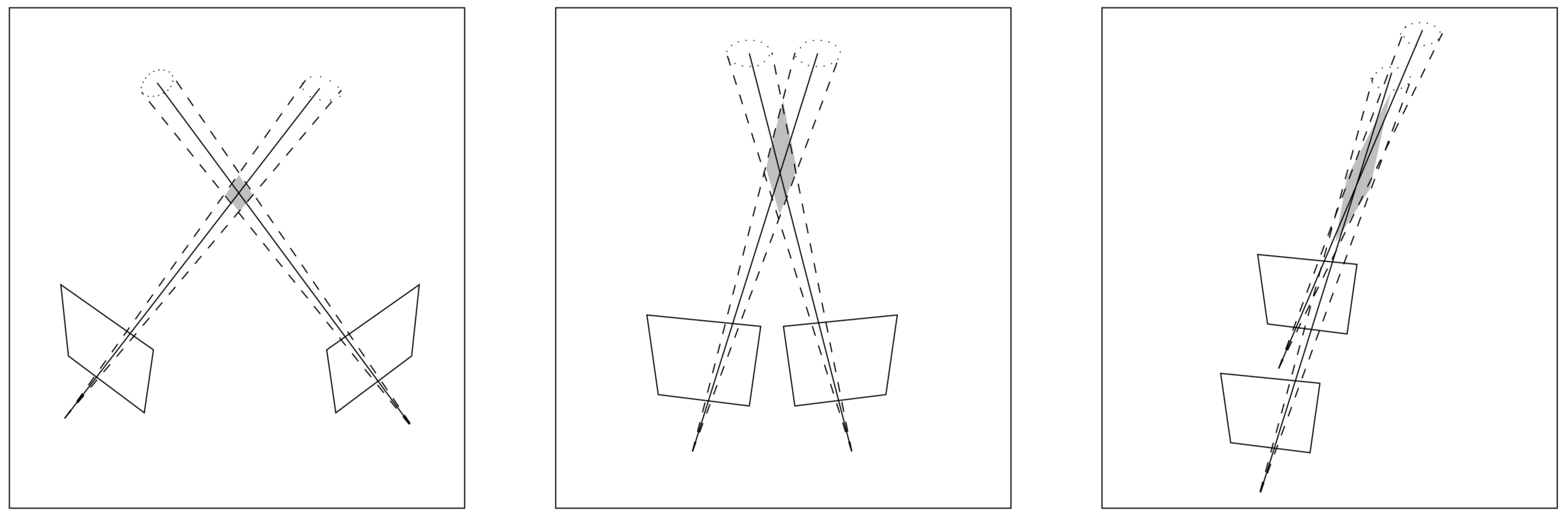}
	\caption{Observability of the world coordinates based on the baseline\cite{Hartley2004}.}
	\label{fig:observability:sfm}
\end{figure}

Figure~\ref{fig:pc_bad} shows the poor observability of the trajectory where the ILS is the center of the field of view over the glide slope. The performance of quadtree and SfM was excellent, but very few of the Monte Carlo iterations met the slope requirements.  Furthermore, traveling a significant distance on a glide slope in the camera boresight direction did not improve the performance.  Analogous to parallax, this relative geometry (translating along the boresight) leads to very poor observability, which leads directly to poor slope estimation. This problem was well presented in the literature by Hartley and Zisserman\cite{Hartley2004}. This problem of observability is illustrated in Figure \ref{fig:observability:sfm}. 

\begin{figure}[H]
	\centering\includegraphics[width=3.5in]{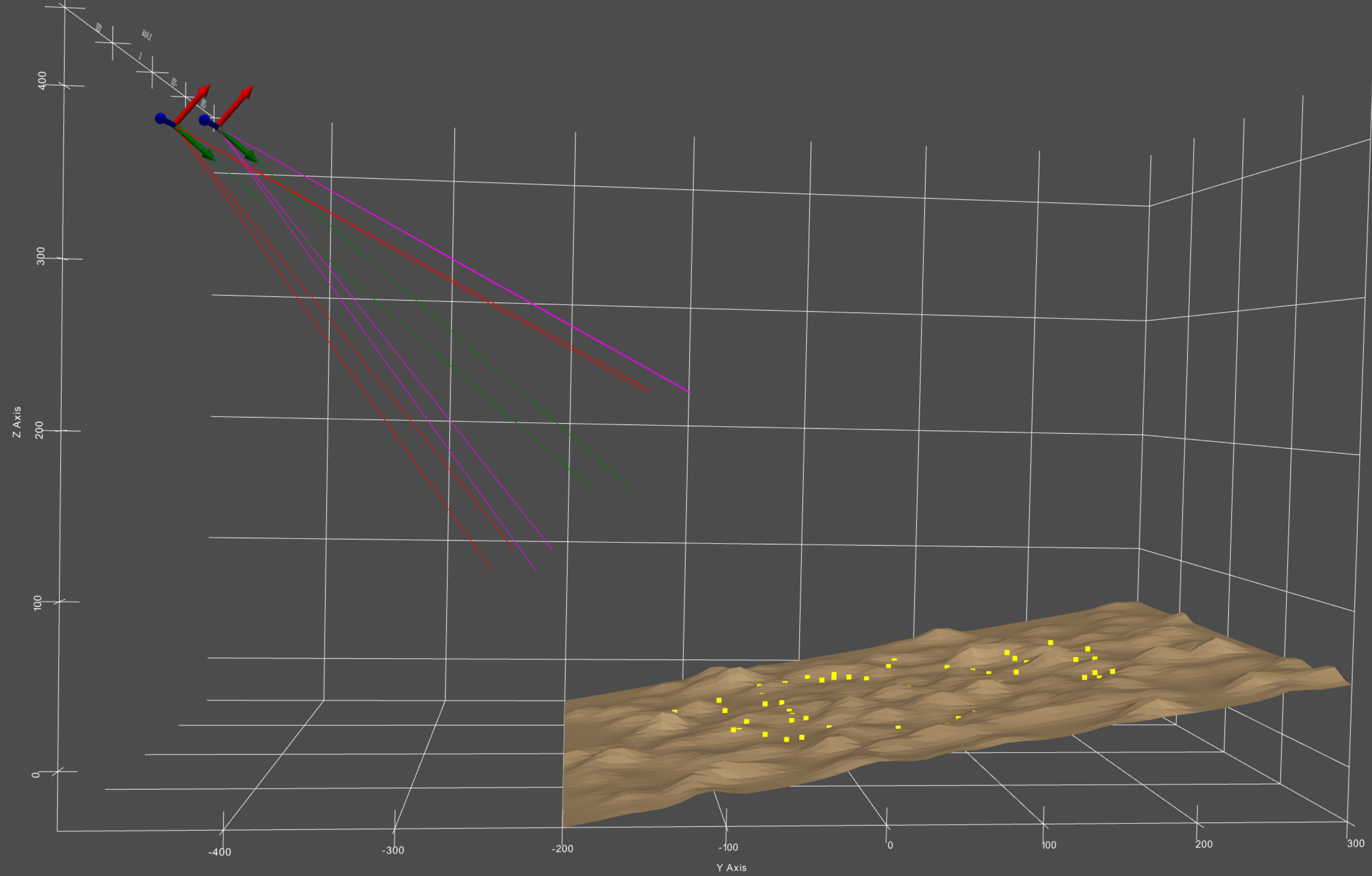}
	\caption{Example surface and random point cloud selection to validate reconstruction and slope.}
	\label{fig:modeled:surface}
\end{figure}

Furthermore, it was determined that a lateral motion with a constant altitude (as shown in Figure~\ref{fig:modeled:surface}) was shown to adequately meet requirements associated with the accuracy of slope computation. The lateral movement, with constant pitch, increased the baseline and produced two overlapping images with improved observability, and therefore reduced error in the slope calculation. 

\begin{figure}[H]
    \centering\includegraphics[width=2.5in]{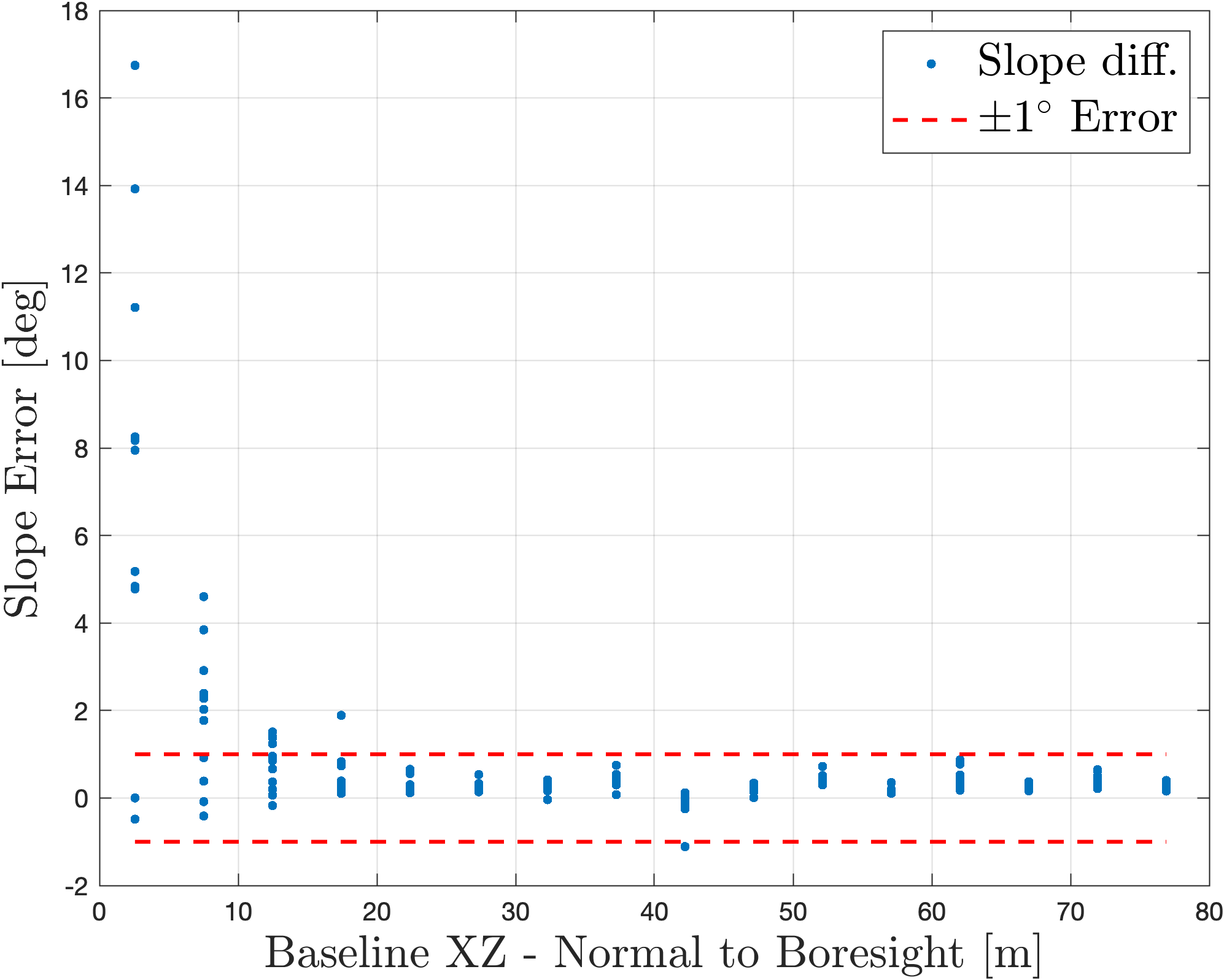}
	\centering\includegraphics[width=2.5in]{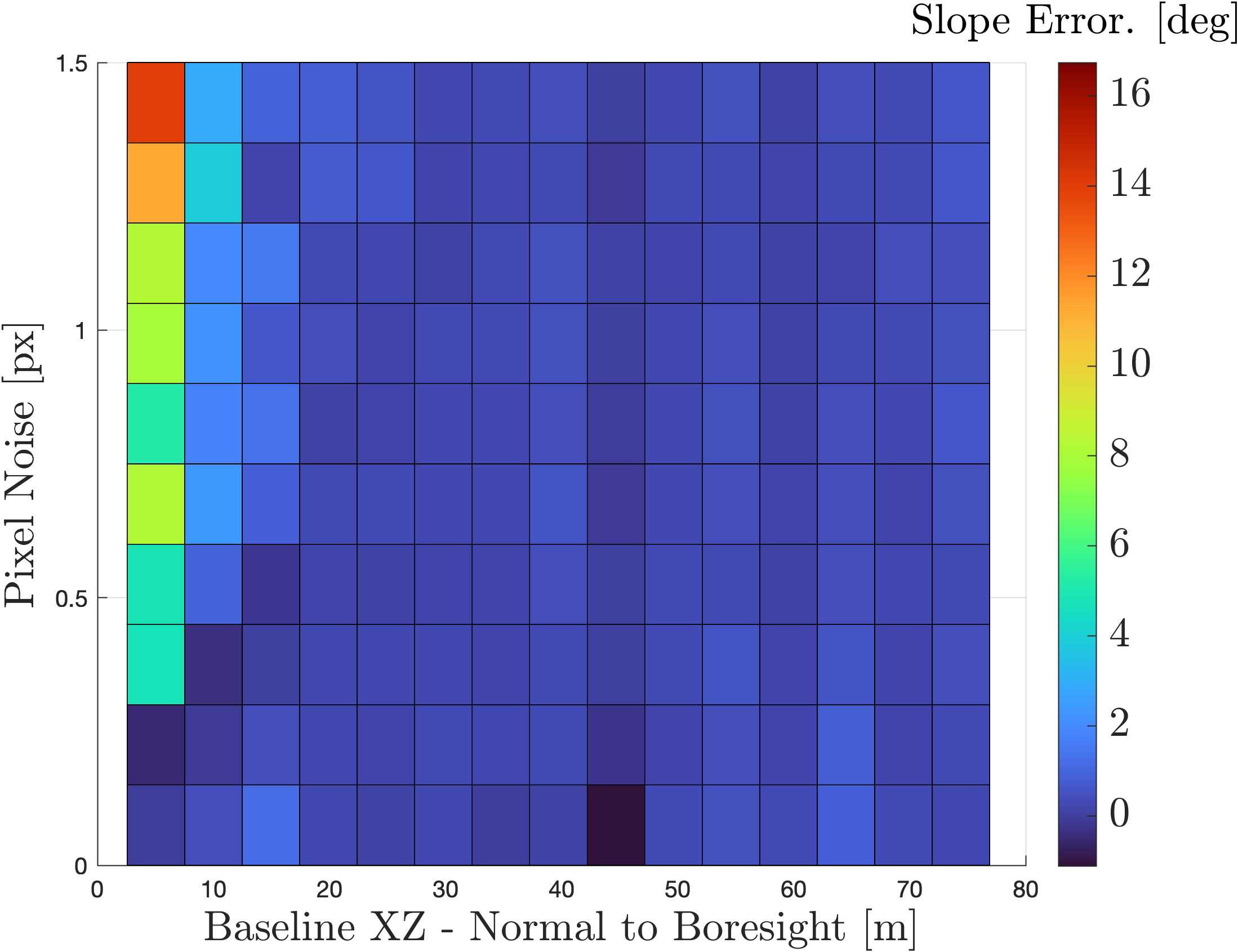}
	\caption{Slope error from the Monte Carlo simulation with synthetic features for lateral relative motion with constant altitude.  Left: Slope error versus change in baseline. Right: Pixel Noise and Slope error versus change in baseline.}
	\label{fig:pc_good}
\end{figure}

Figure~\ref{fig:pc_good} demonstrates that for the Nova-C trajectory, with the navigation hardware baselined, about 30 meters of translation provides the relative motion necessary for point cloud generation that meets the requirements in slope determination.  In the case of NOVA-C's landing trajectory, the 30 meters of translation correlates to  about 2.4 seconds during the period of HDA operations. Figure~\ref{fig:pc_good} was developed under a Monte Carlo test to minimize slope error while simultaneously maximizing observability.


\begin{figure}[H]
	\centering\includegraphics[width=\textwidth]{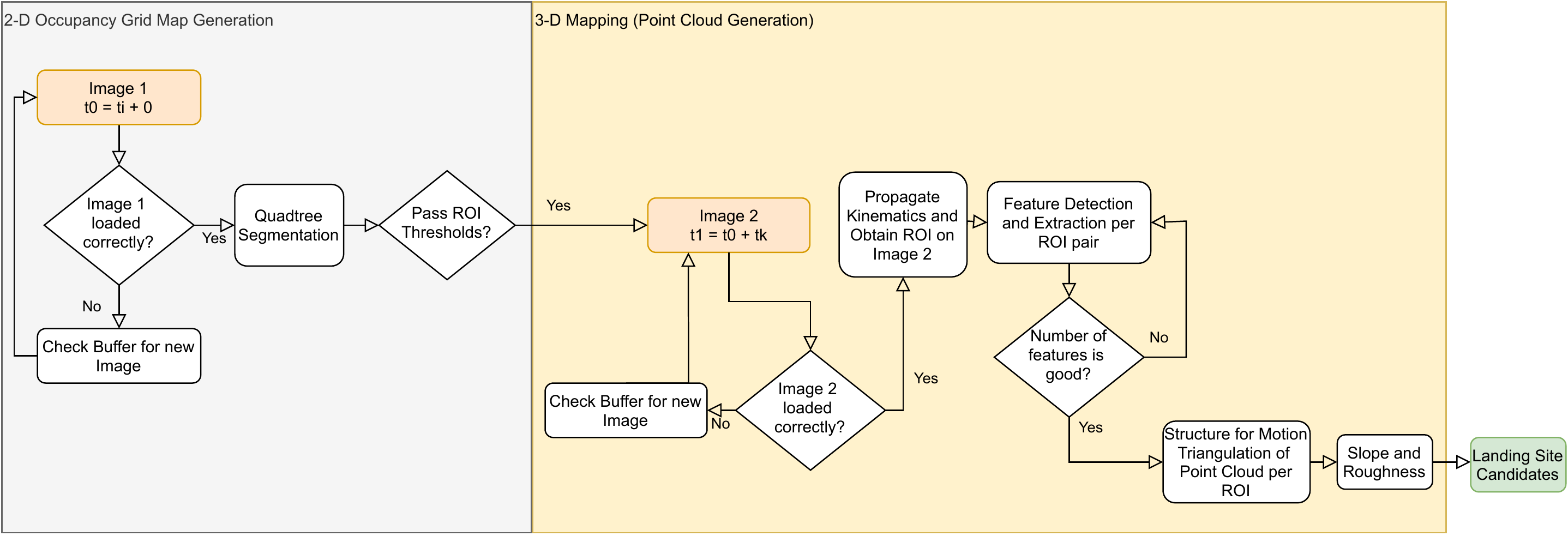}
	\caption{HDA flight software implementation.}
	\label{fig:hda:flowchart}
\end{figure}

Figure~\ref{fig:hda:flowchart} illustrates the implemented flight software algorithm. In summary, it starts by collecting the first image. The quadtree identifies sites with large lighting variations or large areas of shadow, thereby reducing the area required in future processing. Shortly after, a second image is captured, with known (measured by the IMU) relative translation and rotation of the camera between the two images. The locations of the ROIs from quadtree in the first image are predicted in the second image based on the relative motion, again to minimize analysis area. Points and descriptors are identified using ORB in both images, and localized (regional), point clouds are generated. The slopes are then calculated with respect to the local gravity vector. Finally, the ROIs are sorted based on area, slope, and roughness, and then passed to the GNC system. The approach is shown in testing (sections below) to meet the timing and accuracy requirements for safe landing.

\section{Test Campaigns}
\subsection{Overview}
To ensure the success of the HDA system in appropriate landing site selection within computational time constraints, several test campaigns have been and will be performed leading up the launch of the NOVA-C lander.  The campaigns include (but are not limited to):
\begin{itemize}
  \item Software in the Loop (SIL) Campaign.
  \item Truck Test Campaign
  \item Aircraft Testing Campaign
  \item Processor in the Loop (PIL) Campaign.
  \item Helicopter Testing Campaign
\end{itemize}

\begin{figure}[ht]
	\centering\includegraphics[width=2.5in]{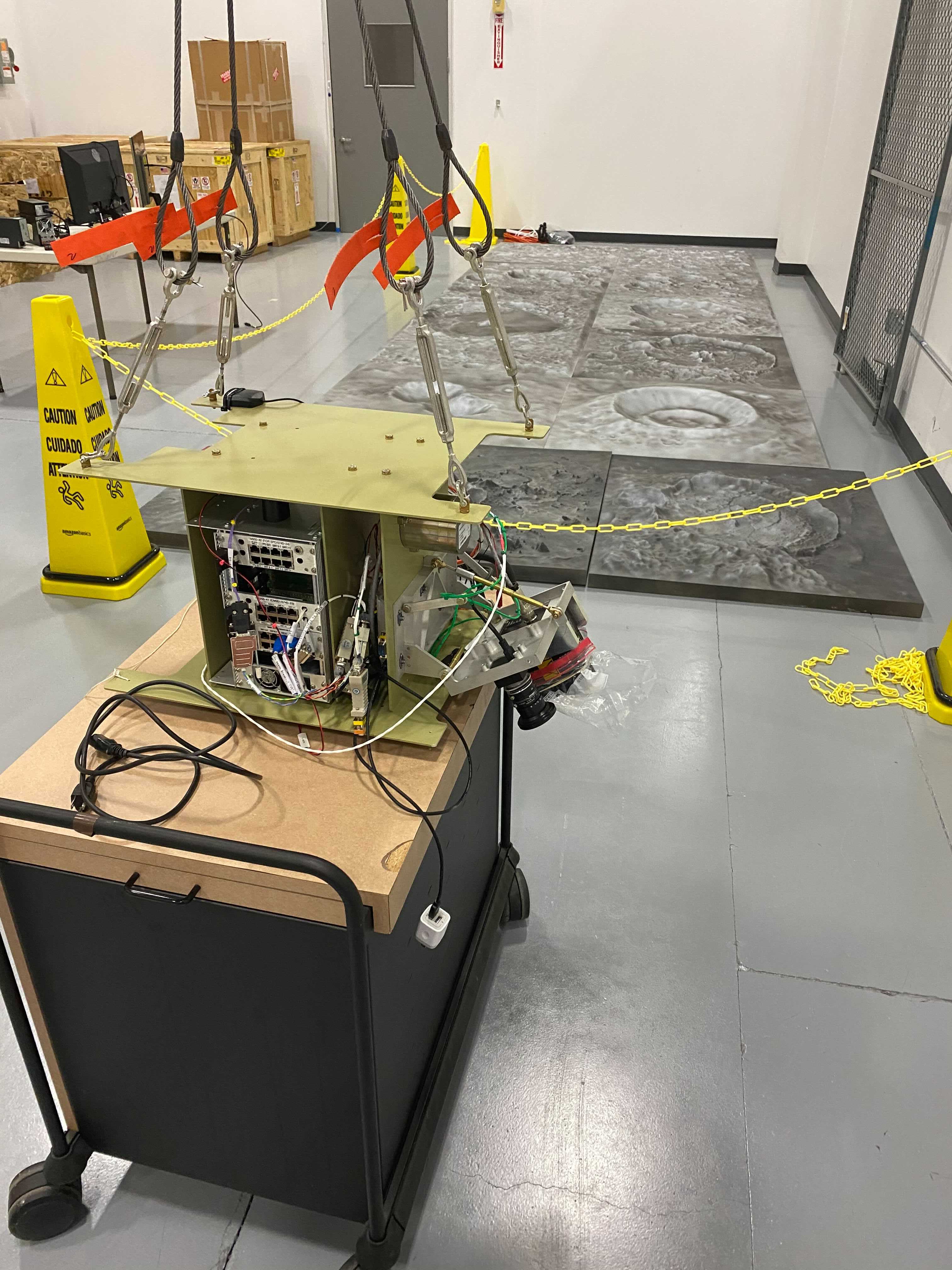}
	\caption{An early iteration of the NOVA-C Navpod including engineering units of IMU, CPU, VPU, optical camera, and laser range finder.}
	\label{fig:navpod}
\end{figure}

As mentioned previously, field test campaigns (truck, aircraft, and helicopter) necessitate a suite of engineering unit sensors and computational platforms. At times, practical and safety considerations have led to Navpod deviations from flight like configurations.  The use of various laser range finders due to eye safety constraints is one example.  Supply chain issues with the flight cameras are another. In the case of the camera, commercial units not suitable for space flight have been utilized as stand ins.  In all cases, care has been taken to ensure performance of the sensors is near what is expected for flight like conditions.  For example, test stand-in cameras have similar field of views and focal lengths as the flight model camera baseline.

\subsection{Software in the Loop Campaign}

It is rather a misnomer to refer the software-in-the-loop testing as a ``campaign", it being more of a continuous and ongoing effort.  This testing involves fusing the flight software code base, a 6 degree-of-freedom simulation, and synthetic imagery to verify, test, and stress the HDA performance.

Because HDA runs open loop to the GNC system until it has completed its assessment of the landing site, extensive testing is possible without in-the-loop synthetic image generation.  The 6 DoF simulation first runs through HDA operation to render synthetic imagery. Then, this synthetic imagery and original simulation outputs are fed into the flight software as a flat file and the HDA algorithms execute.  This workflow allows for HDA test and verification in a very controlled manner.  Additionally, the NOVA-C team is completing work on generating in-the-loop synthetic imagery so that end-to-end testing of the flight system with HDA in the loop and its impacts on landing performance is evaluated extensively.  

\begin{figure}[ht]
	\centering\includegraphics[width=4 in]{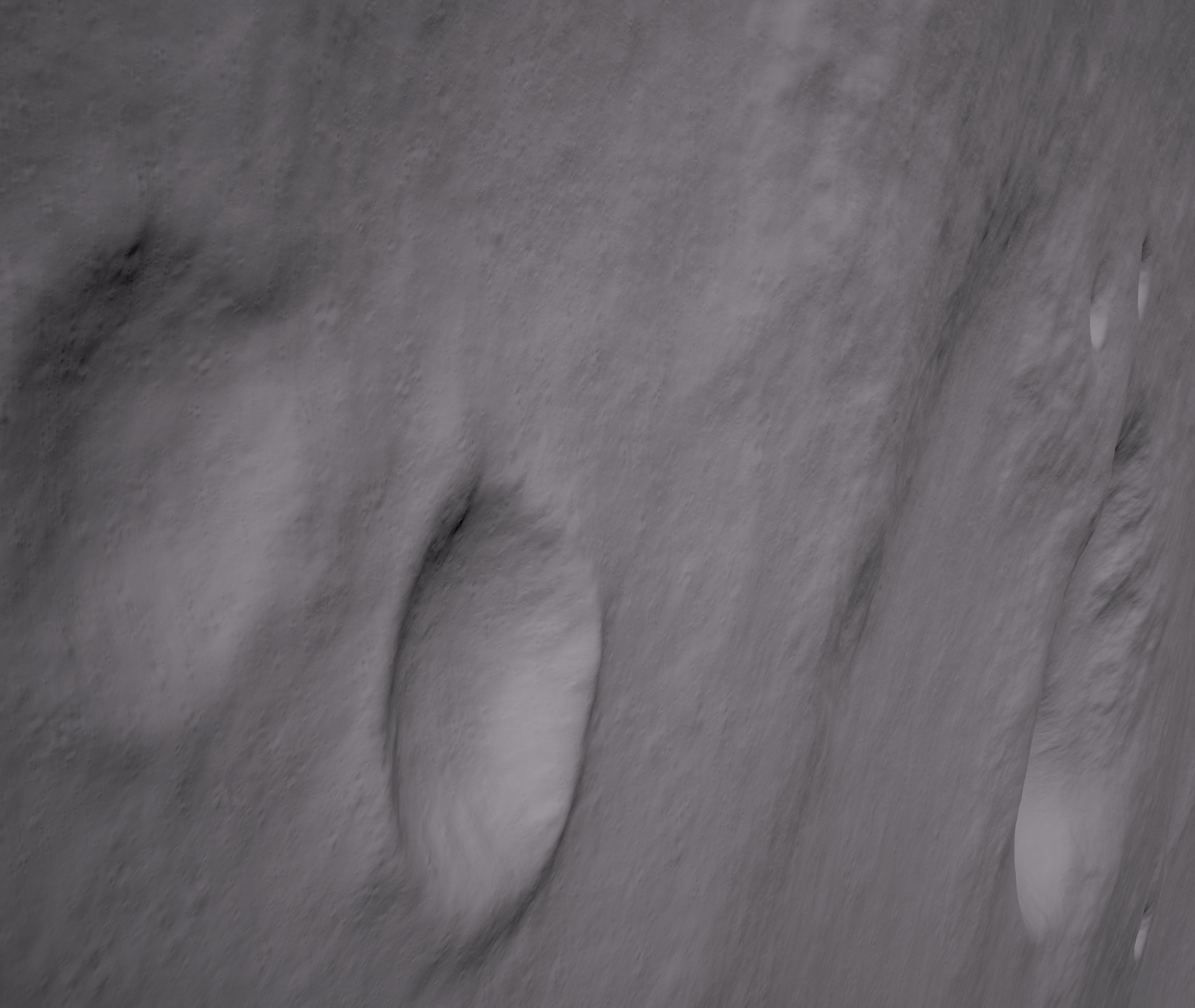}
	\caption{Example of synthetic imagery used to test HDA}
\end{figure}

\subsubsection{Synthetic Imagery}

Synthetic imagery for testing HDA algorithms is built with high resolution synthetic digital elevation maps (DEMs), physically-based materials, artificial hazards, and dynamic lighting. Synthetic DEMs of the lunar surface are built using World Creator, a procedural terrain generator that can produce detailed synthetic landscapes with scattered craters and varying slopes. A 40 x 40 km height map with 0.5 mpp (meters per pixel) resolution is imported into Unreal Engine 4 (UE) as a landscape 3D model and placed in the necessary lunar-centered lunar-fixed (LCLF) location and orientation for rendering. 

Physically based materials are applied to the 3D model using Unreal Engine's material blueprint shader graph to mimic a realistic lunar regolith. The lunar surface material is composed of a three vector RGB base color, a roughness parameter that controls the light reflectivity of the surface, and five layers of seamless normal maps. The normal maps are blended together with appropriate scaling based on the distance between the render camera and the 3D model in order to give the illusion of an extremely high resolution image while using low computational power. At the highest altitude of the HDA image sequence, two normal maps of large craters and rocks are blended and shown while detailed maps are hidden. At the lowest altitude, two normal maps of detailed rocks, pebbles, sand, and dust grain are blended and shown while the larger rocks are hidden. At distances in between, the high altitude and low altitude maps are faded and interpolated with a normal map of medium sized rocks to create a smooth transition in between. The result is a dynamic material that imitates how finer details would come into focus at closer ranges on a lunar surface. 

\begin{figure}[ht]
	\centering\includegraphics[width=6 in]{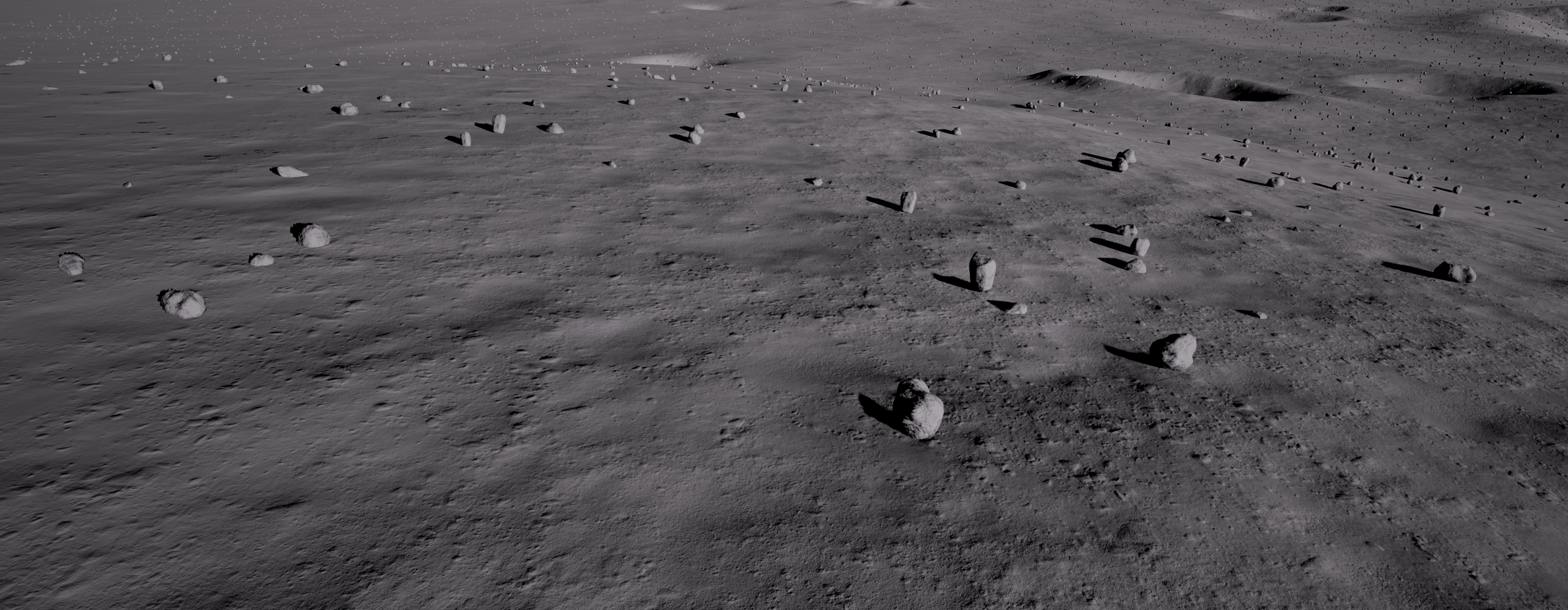}
	\caption{Rock distribution across synthetic terrain}
\end{figure}

A custom procedural hazard generator was created in Unreal Engine to distribute rocks and boulders across any 3D terrain or model. The size and number of rocks scattered in a given square area are based on a rock distribution power law \(N(D)=kD^r\) where the cumulative number of rocks \(N(D)\) is determined with rock diameter \(D\) and the constants \(k\) and \(r\) which are based on a specific location on the moon.\cite{LI201730} When the procedural generator is provided an area's radius, rock diameter minimum, rock diameter maximum, and constants k and r, it determines how many rocks of a certain size should be distributed within the radius. Since a high volume of 3D models and polygons in any scene can become too computationally heavy, instancing of eight unique rock meshes allows the engine to produce thousands of rock instances across the terrain. The hazards' locations and rotations are randomized by a given seed value, and can quickly iterate through unique distributions with the ability for cases to be reused. 

The camera and sun LCLF positions and rotations are provided by the mission design tool and used to establish the lighting and rendering. A directional light component creates dynamic lighting and cast shadows across the terrain. Camera parameters, such as horizontal field of view, aspect ratio, exposure, and resolution are based on estimated values of the on-board camera. 

Ground truth data for HDA algorithms (e.g., point clouds) are available through custom functions in Unreal Engine's blueprints. To obtain a point cloud of the terrain section captured in a single render, the function positions the camera in the same location and rotation as when the render was created and iterates through a given number of points by column and row. A ray from each point is traced from the camera to the terrain until it returns a hit location in LCLF. Each point's location is saved to an output CSV file which permits other applications to read the data. 
\begin{figure}[ht]
  \centering
  \begin{minipage}[b]{0.4\textwidth}
    \includegraphics[width=\textwidth]{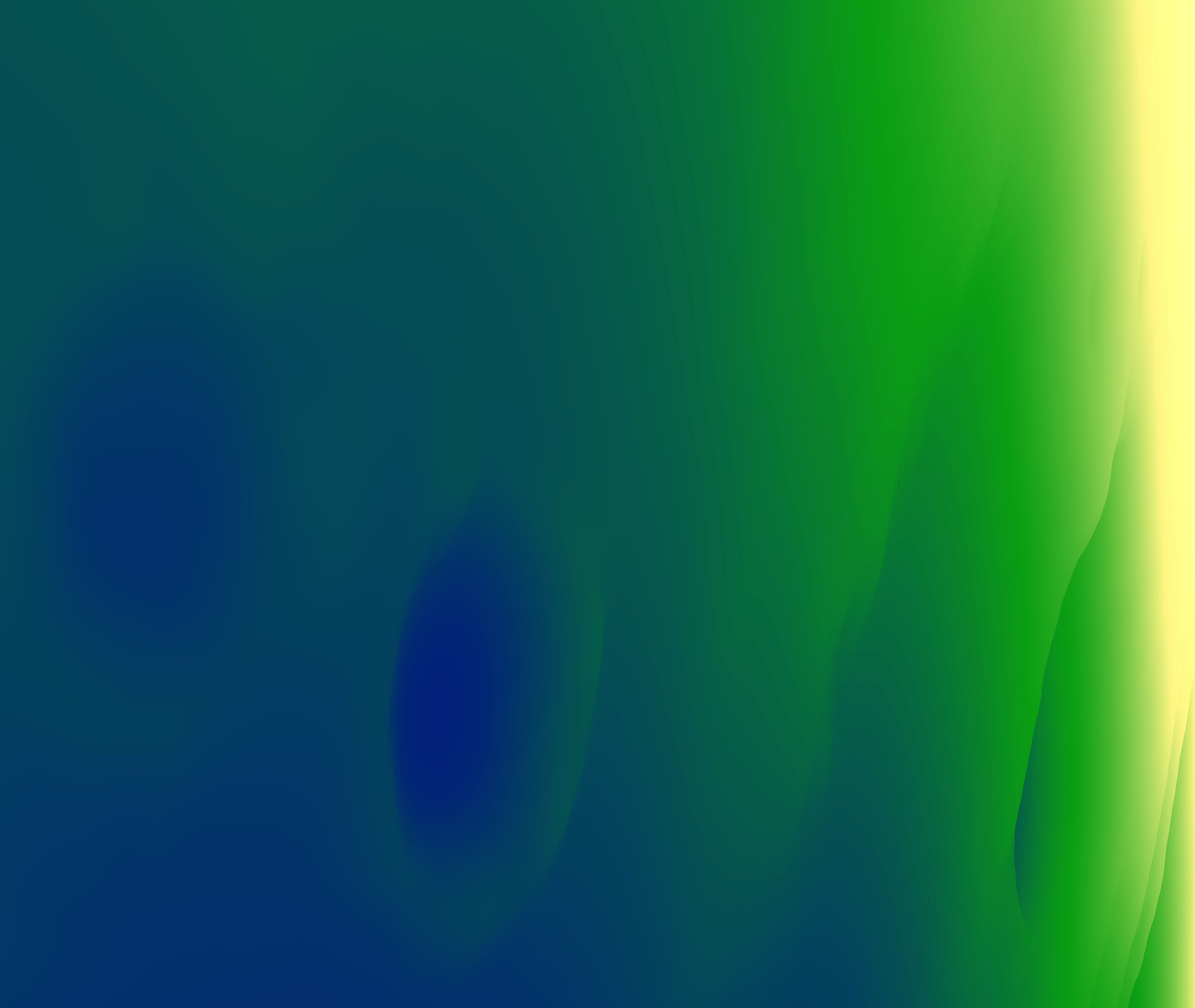}
    \caption{Elevation visualization map}
  \end{minipage}
  \hfill
  \begin{minipage}[b]{0.4\textwidth}
    \includegraphics[width=\textwidth]{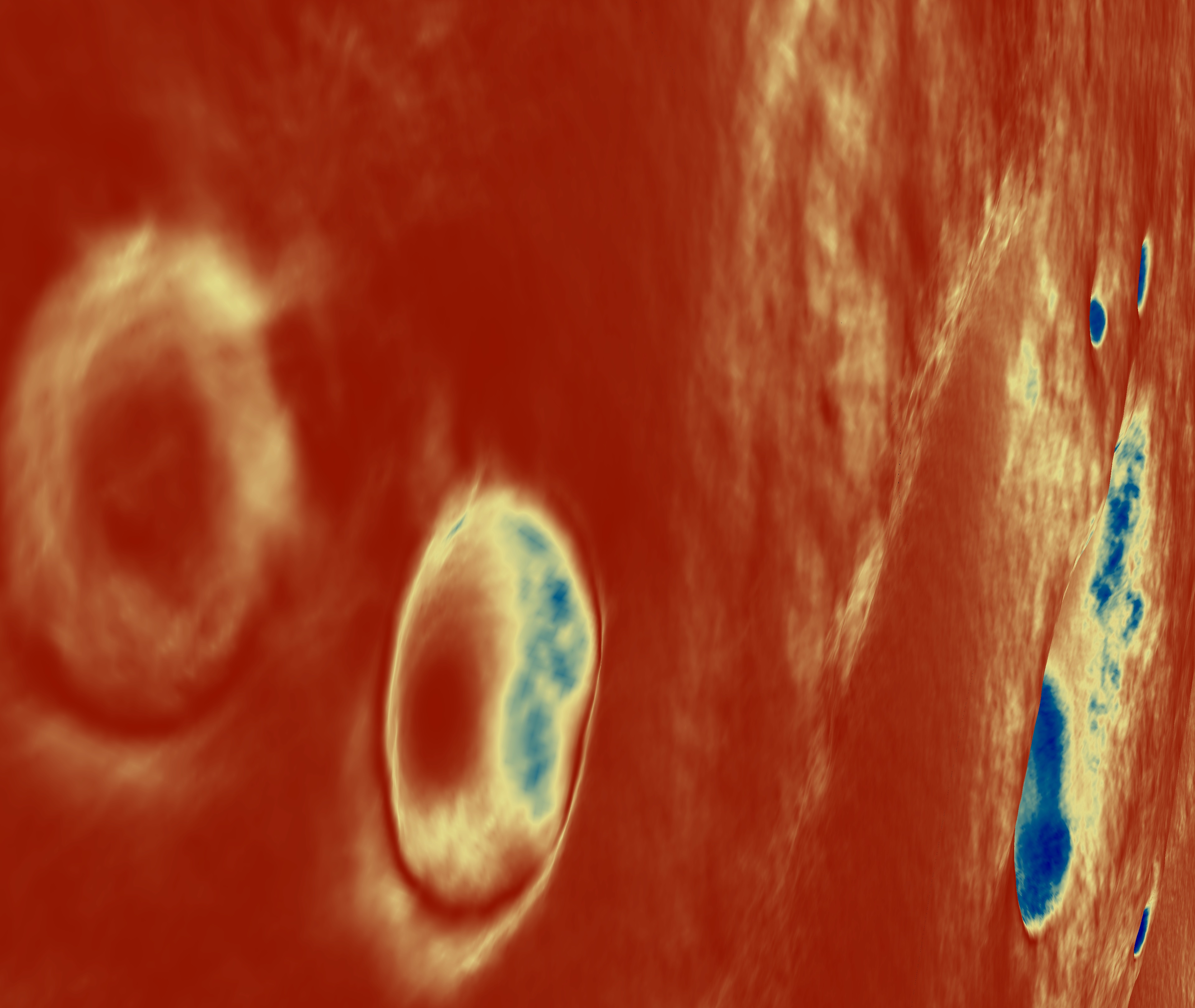}
    \caption{Slope visualization map}
  \end{minipage}
\end{figure}

Terrain color maps that indicate the elevation and slope of the terrain in a render are depicted through visualization materials. To visualize the elevation of a terrain in LCLF position, the absolute world position of each vertex on the terrain is remapped from a minimum and maximum elevation to three RGB colors to create a color map. Visualizing the slope of a terrain uses a similar logic and remaps the minimum and maximum slope of a vertex's normal direction in degrees to three RGB colors. Since lighting and shadows of a normal render can affect the true color of the base material, the color map is strictly rendered using a base color pass. 

\subsubsection{SIL Testing Results}
The HDA SIL system fuses a combination of images and navigation tables to perform the main tasks of hazard detection and hazard avoidance. The images are generally those synthetic images produced in the Unreal pipeline discussed earlier in a previous section, but this is not necessarily the case. After each run of truck or aircraft testing, the navigation data and images are packaged and converted to the same format as the synthetic data and images. In this way we performed more than one run on each set of data obtained from physical testing and ensure performance and accuracy met specifications and expectations, albeit on different hardware. 

\begin{figure}[ht]
  \centering
  \includegraphics[width=0.5\textwidth]{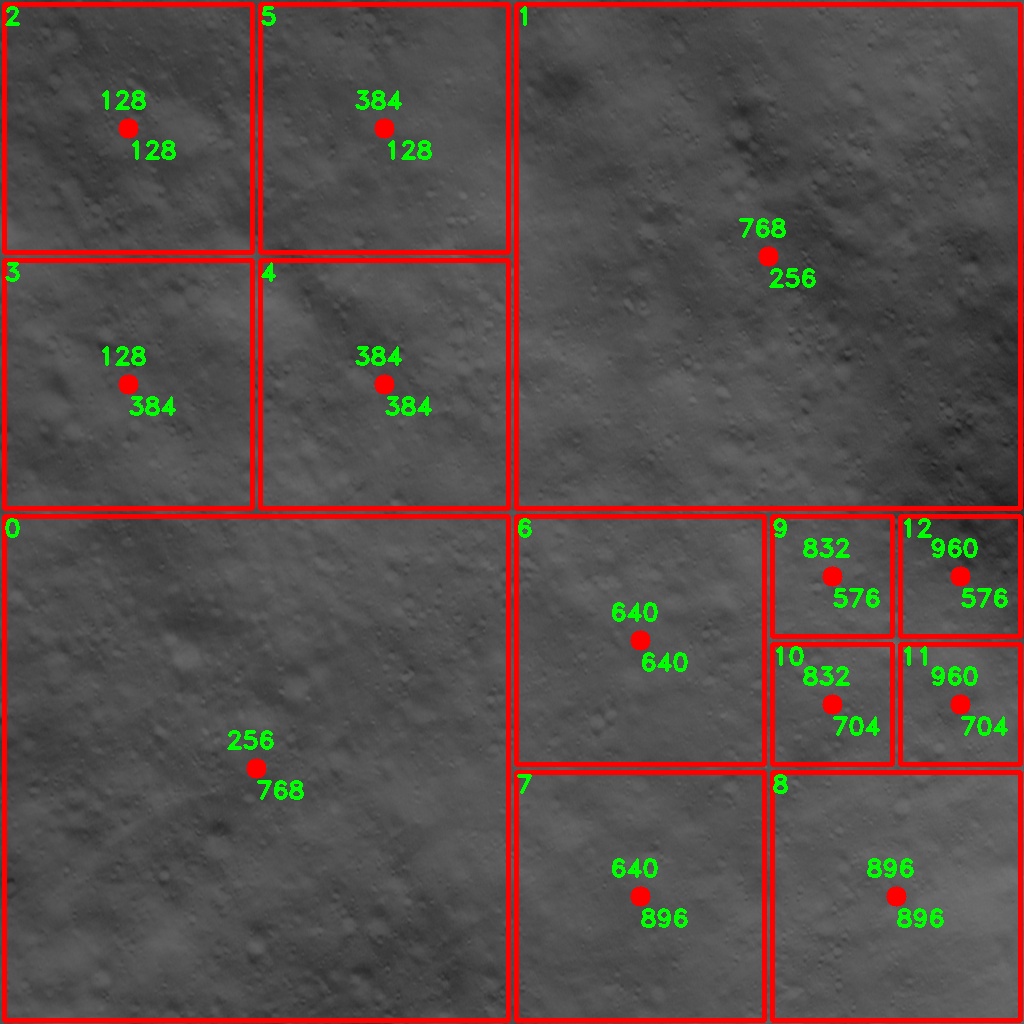}
  \caption{Example quadtree using synthetic images. The green numbers at the center of each ROI represent the starting pixel values of the original image, and the numbers at the top left of each is the index number. Note that quadtree does not necessarily fill all areas of the cropped image with a valid ROI which we show in a later figure. }
  \label{fig:quadtreeSynthetic}
\end{figure}

\begin{figure}[ht]
  \centering
  \includegraphics[width=0.925\textwidth]{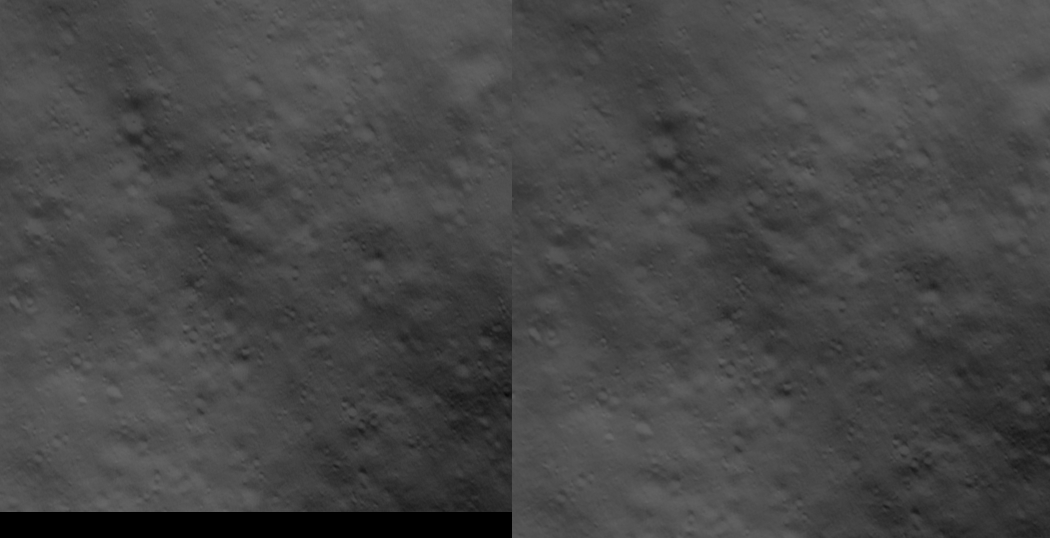}
  \caption{Two chosen ROIs for first and second images showing roughly the same region. Note that the second ROI is larger to account for navigation error.}
  \label{fig:roiSpace}
\end{figure}

\begin{figure}[ht]
  \centering
  \begin{minipage}[b]{0.48\textwidth}
    \includegraphics[width=\textwidth]{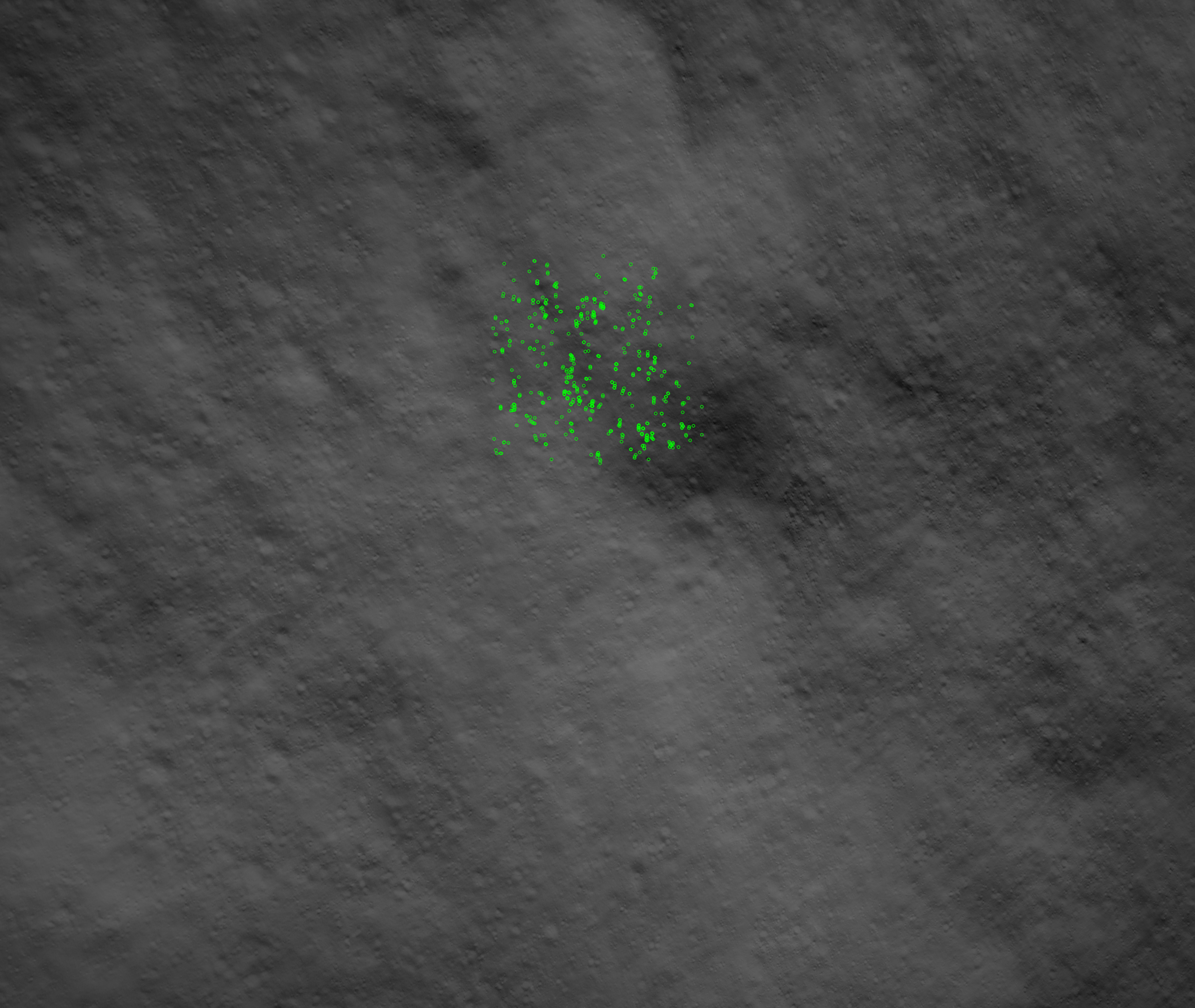}
  \end{minipage}
  \hfill
  \begin{minipage}[b]{0.48\textwidth}
    \includegraphics[width=\textwidth]{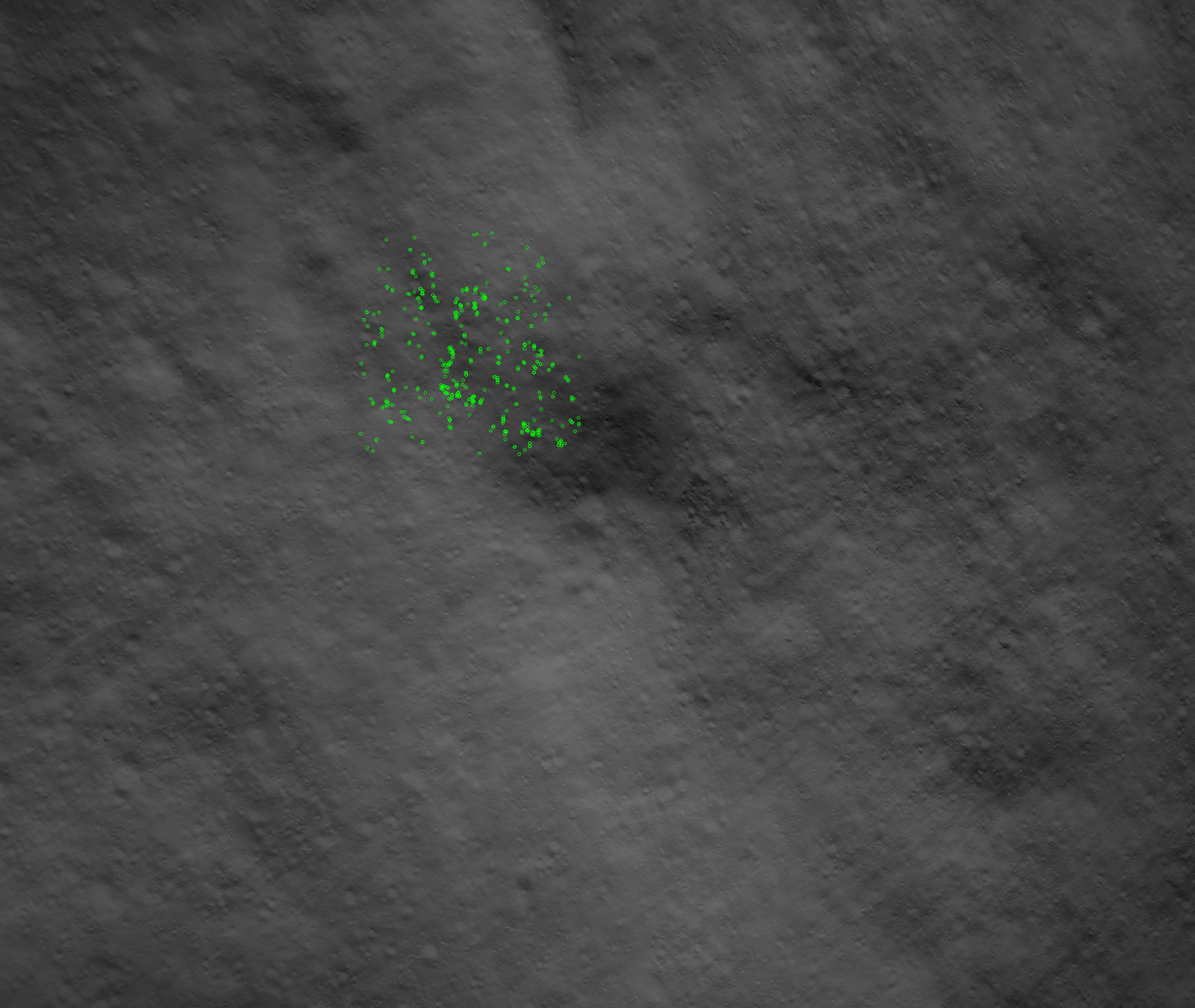}
  \end{minipage}
\caption{Keypoints for a given ROI in green for first HDA image (left) and second image (right). These are the same keypoints in the image above showing just the ROIs. These keypoints are those keypoints that have survived both the initial RANSAC rejection and the nav-based outlier rejection.}
\label{fig:keypointsSIL}
\end{figure}

Below is an example of the results from such tests, contained in Figures~\ref{fig:quadtreeSynthetic}, \ref{fig:roiSpace}, and \ref{fig:keypointsSIL}. We show the results from both the quadtree and SfM algorithms. The synthetic imagery was loaded into the simulation and the navigation information along the trajectory is loaded into an input file simulating how such data will be available to the HDA algorithm. This information includes transformations from different frames and sensor data about the simulated vehicle state. With both images and navigation, we can complete a simulated run of HDA. The first step is breaking down the image into smaller, more uniform regions of interest (ROI) using the quadtree algorithm. This algorithm uses recursion to break down an area into four equal-sized quadrants and continues until either the given ROI meets certain criteria or the subdivision sizes fall below a given threshold.

\begin{figure}[h!]
  \centering
  \includegraphics[width=1\textwidth]{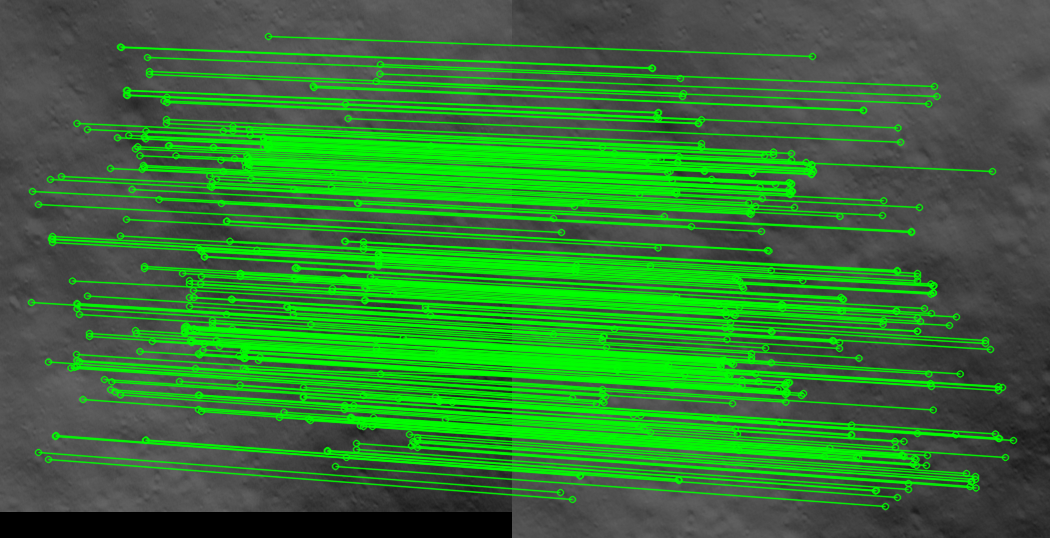}
  \caption{Matches between keypoints in the first and second images, shown in green.}
  \label{fig:matchesSIL}
\end{figure}

Once the quadtree is complete, the algorithm waits until enough time has passes until it resumes with structure-from-motion from two images. Using OpenCV's ORB detector, we pass in the ROI from the first image and the corresponding ROI from the second image, which we calculate from the navigation data and enlarge the second ROI to account for uncertainty. After applying an outliers rejection method, the algorithm computes the reconstructed 3D point cloud as well as the corresponding slope and roughness. Figures~\ref{fig:roiSpace}, \ref{fig:keypointsSIL}, and \ref{fig:matchesSIL} illustrate the results for a standard synthetic simulation run.\\

Similarly, based on the results from completed testing regimes, we also processed the test runs through this same pipeline by fusing the images and navigation data into the same format used for a fully synthetic run. Results here are denoted as hybrid simulations. We have included the hybrid simulation results in later sections.

The accuracy of the slope calculation is critical to mission safety, so using the known point cloud of the lunar surface in the images we compare that to the calculated slopes from the HDA output. The results in Table~\ref{table:sil}
\begin{table}[h!]
\caption{Simulation Slope Calculation Results} 
\centering 
\resizebox{6cm}{!} 
{ 
\begin{tabular}{c c c c} 
\hline\hline 
ROI & Truth Slope & Est. Slope & \# Points \\ [0.5ex] 
\hline 
0  & $5.91^{\circ}$ & $6.30^{\circ}$ & 46 \\ 
1  & $3.64^{\circ}$ & $3.89^{\circ}$ & 46 \\ 
2  & $1.74^{\circ}$ & $1.64^{\circ}$ & 71 \\
3  & $2.56^{\circ}$ & $2.21^{\circ}$ & 84 \\
7  & $6.41^{\circ}$ & $9.13^{\circ}$ & 15 \\
9  & $3.23^{\circ}$ & $8.22^{\circ}$ & 15 \\
11 & $5.39^{\circ}$ & $6.48^{\circ}$ & 26 \\
12 & $2.03^{\circ}$ & $2.55^{\circ}$ & 22 \\ [1ex] 
\hline 
\end{tabular}
}
\label{table:sil} 
\end{table}
%

\subsection{Truck Campaign}
\begin{figure}[H]
  \centering
  \includegraphics[width=0.7\textwidth]{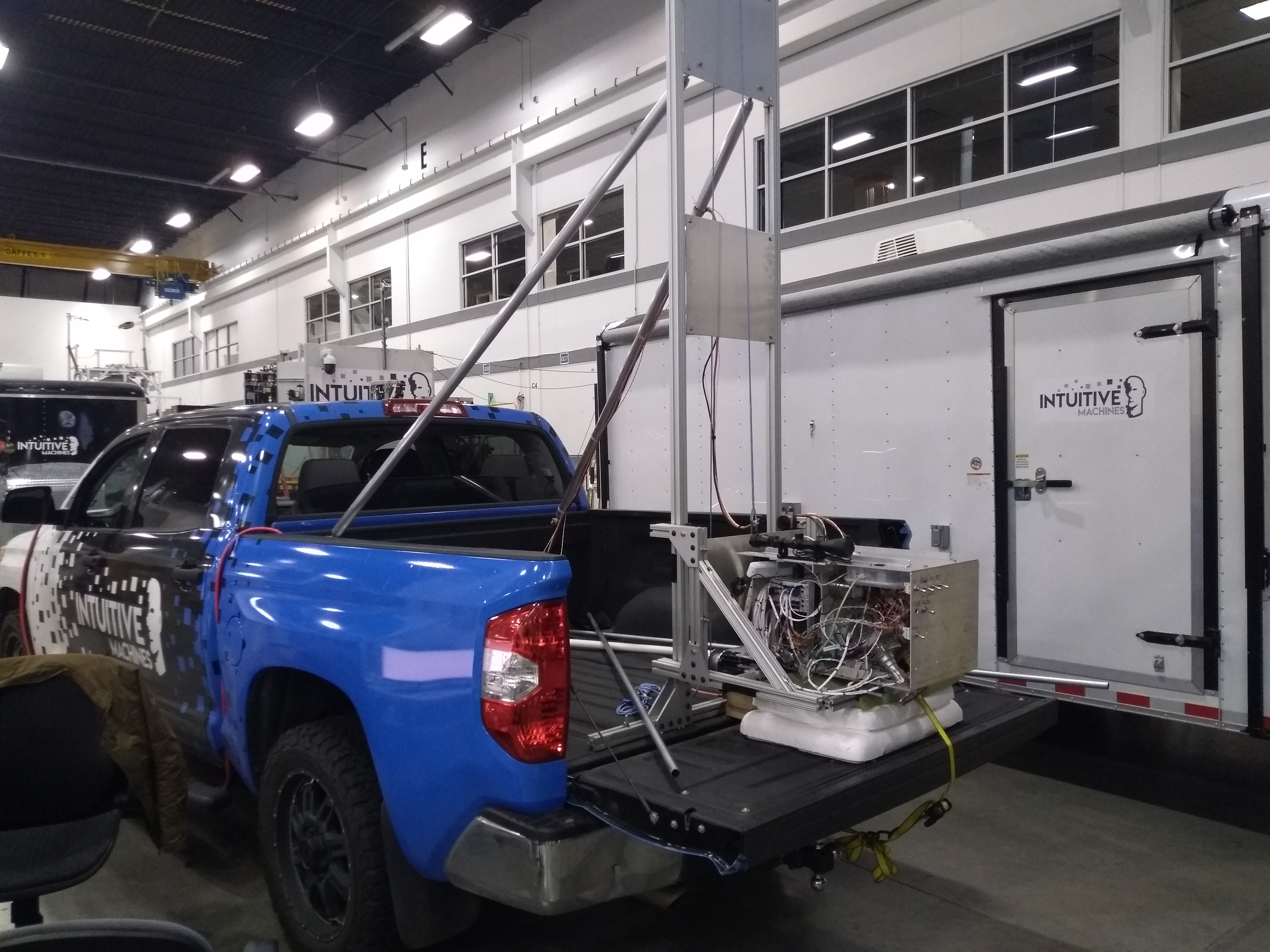}
  \caption{NOVA-C Navpod mounted to the bed of a pickup truck.  A motorized lift provides some vertical motion.  The camera and laser ranger point out the rear at a 45 degree angle.}
  \label{fig:navpod_truck}
\end{figure}

A necessary and empowering component of the test campaign, is the ability to quickly test HDA algorithms with hardware in-the-loop in a real world environment. Mounting the Navpod on the bed of a pickup truck with a motorized lift allows the NOVA-C team to test the HDA algorithms and troubleshoot real-time implementation.  Figure~\ref{fig:navpod_truck} shows the truck with the Navpod mounted.

Because of eye safety concerns, the engineering unit of the laser range finder was replaced with a commercial laser ranging unit.  An additional concern for testing of in this configuration is scale of the problem.  Here, the Navpod rests 1 - 3 meters above the observed terrain vs. the planned 400 meters in altitude for the lunar mission.  Parameter updates to quadtree to segment to the appropriate scale is one of many examples of configurations updates needed to ensure a reasonable test.  Trajectory considerations such that image overlap for the structure for motion is similar to that of what is expected at the Moon is another.  Finally, noise bias from the Navpod IMU accelerometer needs to be observed to ensure that the navigation estimate of the relative motion matches the collected imagery.

\begin{figure}[H]
  \centering
  \includegraphics[width=0.5\textwidth]{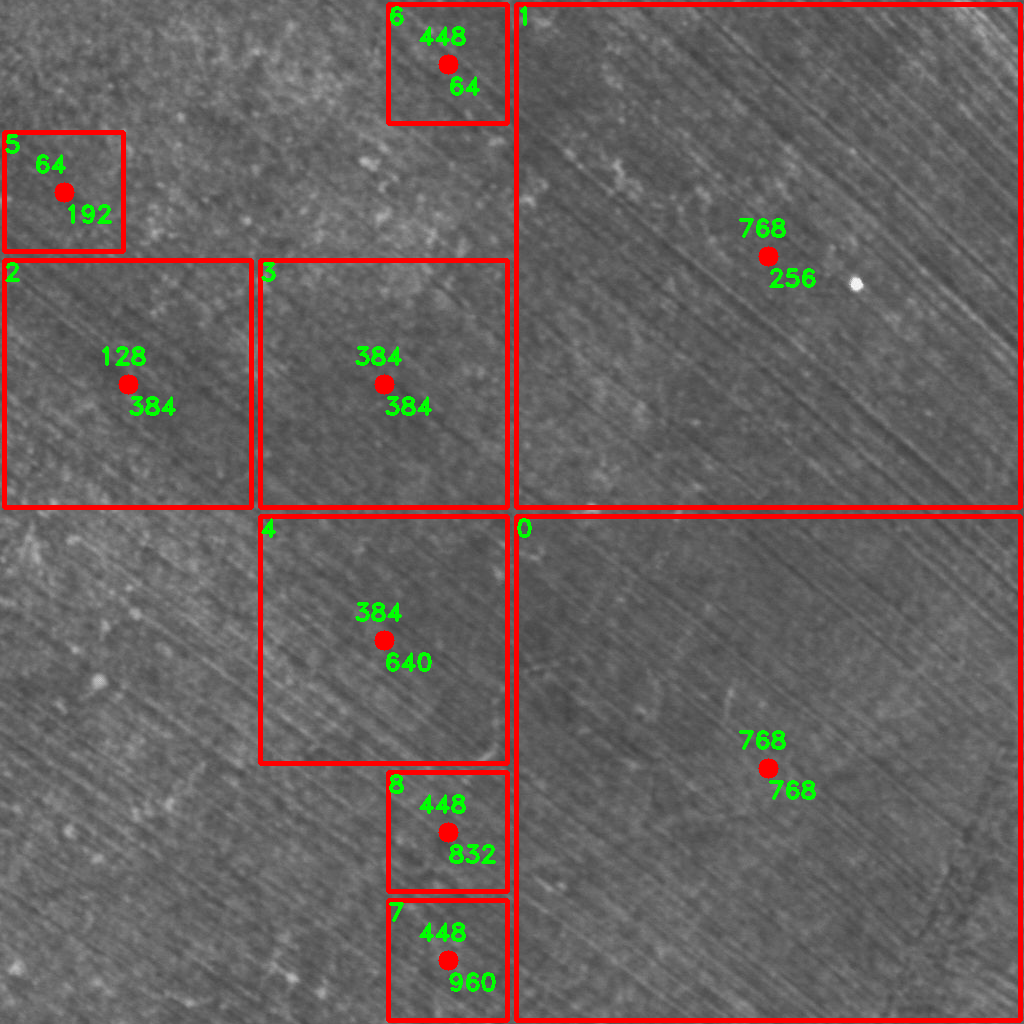}
  \caption{Example quadtree using image from truck testing; the image is of the parking lot concrete from a height of about 4 meters. Note that, in contrast to the previous quadtree result, this quadtree result rejects certain portions of the image due to high standard deviation regardless of how much it recurses.}
  \label{fig:quadtree_truck}
\end{figure}

Just as before, the first step was to break down the image in manageable pieces of the images; the quadtree result from a truck test is included in Figure~\ref{fig:quadtree_truck}. The breakdown of the whole image results in some sections lacking an ROI since each of those regions failed to satisfy the criteria for a candidate landing site. As Figure~\ref{fig:quadtree_truck} shows, the more uniform sections form much larger ROIs, and as the ``roughness'' of the section increases, the size decreases until a section is deemed invalid due to size criteria.
\begin{figure}[H]
  \centering
  {\color{white}
\stackinset{l}{0.5\textwidth}{b}{.0in}{\rotatebox{90}{\rule{3.0in}{1pt}}}{%
  \includegraphics[width=1.0\textwidth]{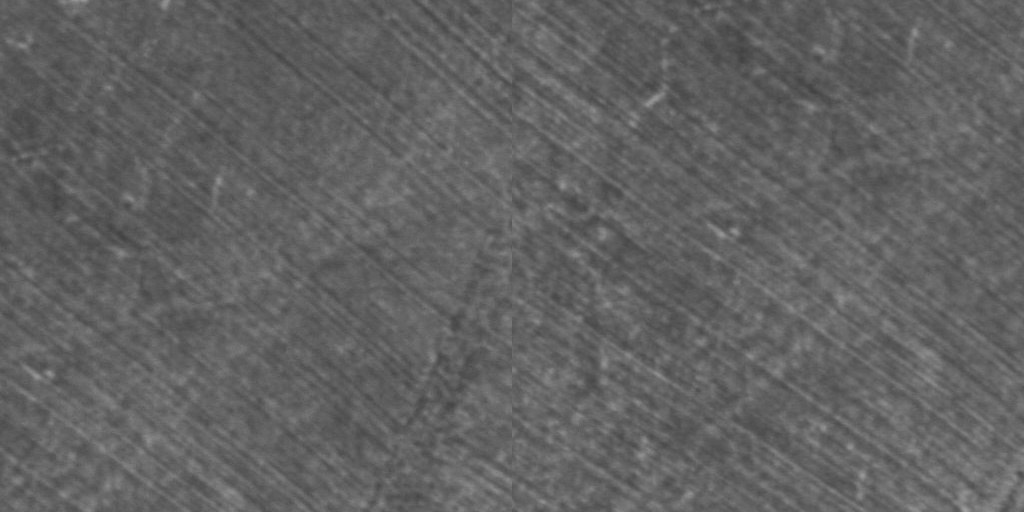}%
}}
  \caption{Two chosen ROIs for first and second images showing roughly the same region. Note that the second ROI is larger to account for navigation error.}
  \label{fig:roi_space_truck}
\end{figure}
\begin{figure}[H]
  \centering
  \begin{minipage}[b]{0.45\textwidth}
    \includegraphics[width=\textwidth]{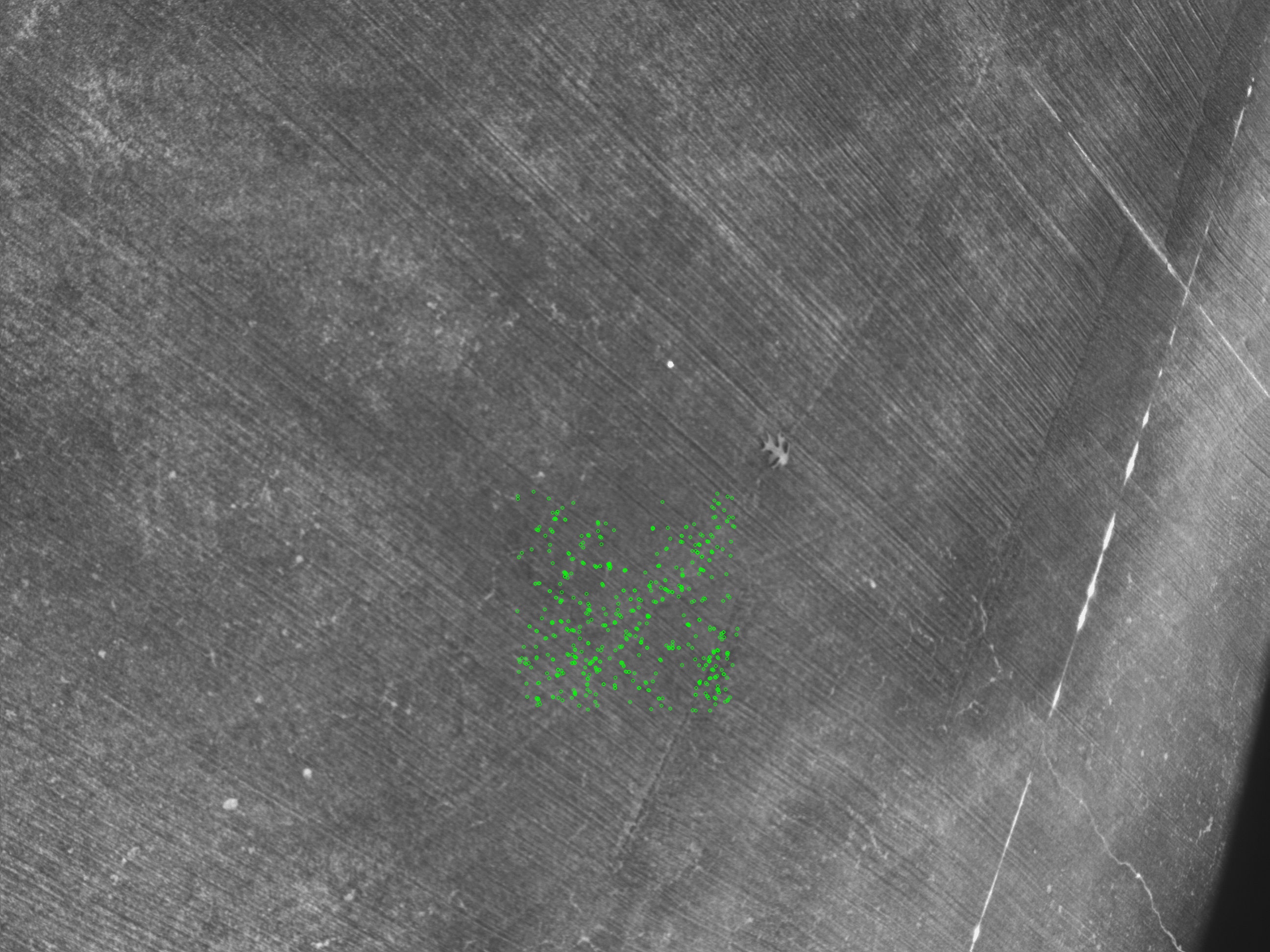}
  \end{minipage}
  \hfill
  \begin{minipage}[b]{0.45\textwidth}
    \includegraphics[width=\textwidth]{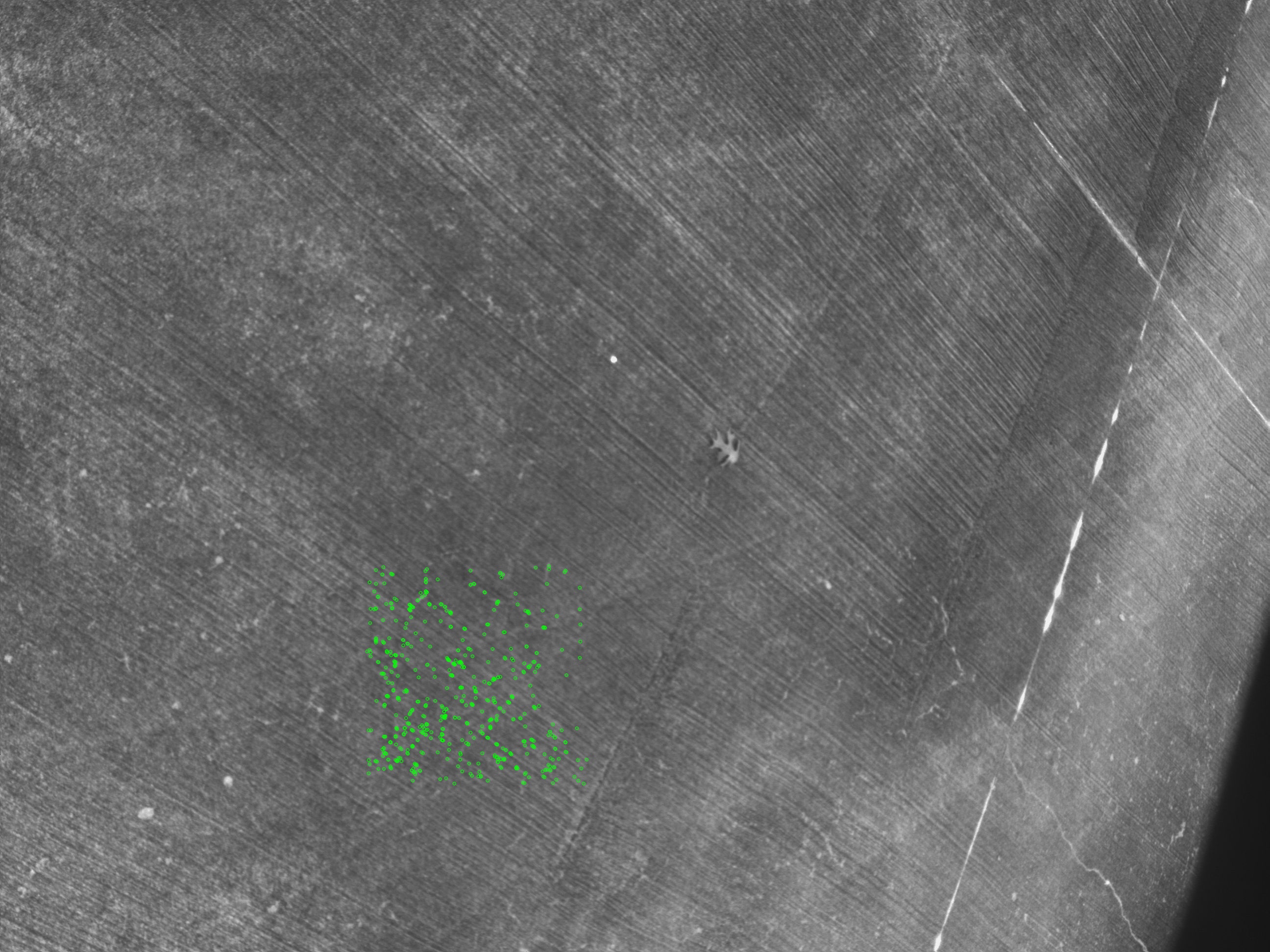}
  \end{minipage}
  \label{fig:keypoints}
\caption{Keypoints for a given ROI in green for first HDA image (left) and second image (right). These are the same keypoints in the image above showing just the ROIs. These keypoints are those keypoints that have survived the initial RANSAC rejection but not the outlier rejection based on navigation data. Note the obvious outlier in the data at a noticeable angle from the others.}
\end{figure}
\begin{figure}[H]
  \centering
  \includegraphics[width=0.7\textwidth]{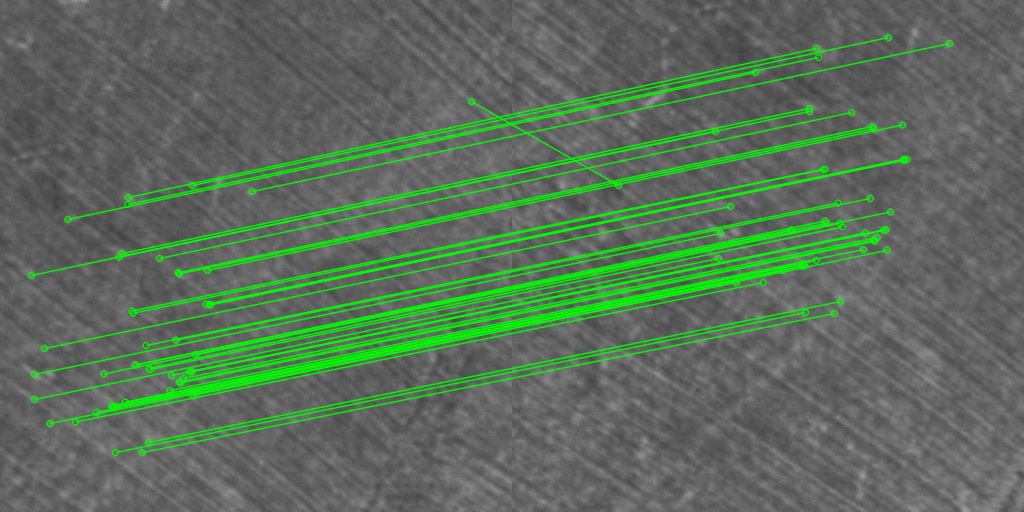}
  \caption{Matches between keypoints in the first and second truck images, shown in green.}
  \label{fig:truck_matches}
\end{figure}
%

\subsection{Aircraft Testing Campaign}

Aircraft testing was performed in the fall of 2021 at the White Sands Missile Range in Las Cruces, New Mexico.  Since HDA was a secondary test objective in favor of the larger navigation system, limited opportunities for HDA specific testing occurred over the one week test period.  Nevertheless, as the aircraft flew constant altitude and straight line trajectories, which is analogous to the planned lunar trajectory for HDA operations, valuable HDA testing and data collection was feasible.

\begin{figure}[H]
	\centering\includegraphics[width=3.5in]{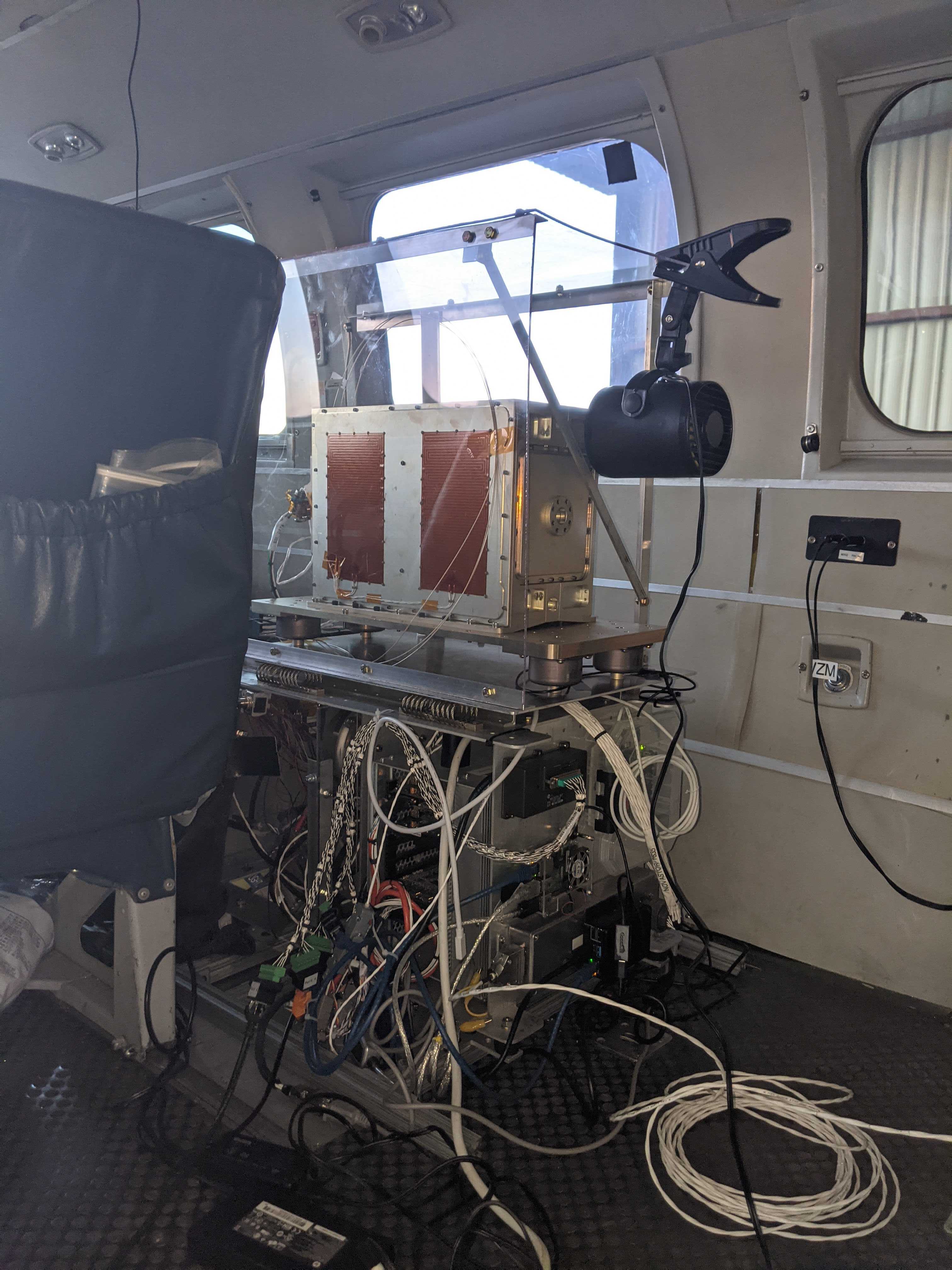}
	\caption{Navpod mounting for aircraft testing.}
	\label{fig:navpod_aircraft}
\end{figure}

Figure~\ref{fig:navpod_aircraft} shows the mounting of the Navpod in the aircraft.  A cutaway in the aircraft belly allowed for an engineering unit of the laser range finder and a visual camera to observe the terrain.  Both were mounted 45 degrees off of nadir towards the nose of the aircraft, once again simulating flight like geometries.  Trajectories generally involved the aircraft traveling at a ground speed of about 80 m/s over various altitudes (most commonly around 1800 meters).  This deviates from the expected lunar HDA operations ground speed and altitude (about 13 m/s and 400 meters) so care was taken to scale relevant HDA parameters, such as the time step between images, in order to get representative performance.

Unfortunately, real-time HDA results were not generated due to a systematic issue with the Navpod's attitude reference and horizon sensor (AHRS).  This particular AHRS sensor depends on the magnetic field for heading fixes and the noisy electromagnetic environment inside the aircraft was not concussive to needed attitude requirements.

Nevertheless, the HDA team utilized the real-world imagery from the aircraft testing campaign to generate offline solutions and evaluate performance using a fusion of the images and navigation and sensor data in Python, which was then transferred to the FSW simulation pipeline.

Just as before, the first step was to break down the image in manageable pieces of the images; the quadtree result from a truck test is included in Figure~\ref{fig:quadtree_truck}. The breakdown of the whole image results in some sections lacking an ROI since each of those regions failed to satisfy the criteria for a candidate landing site. As Figure~\ref{fig:quadtree_truck} shows, the more uniform sections form much larger ROIs, and as the ``roughness'' of the section increases, the size decreases until a section is deemed invalid due to size criteria.
\begin{figure}[H]
  \centering
  \includegraphics[width=0.5\textwidth]{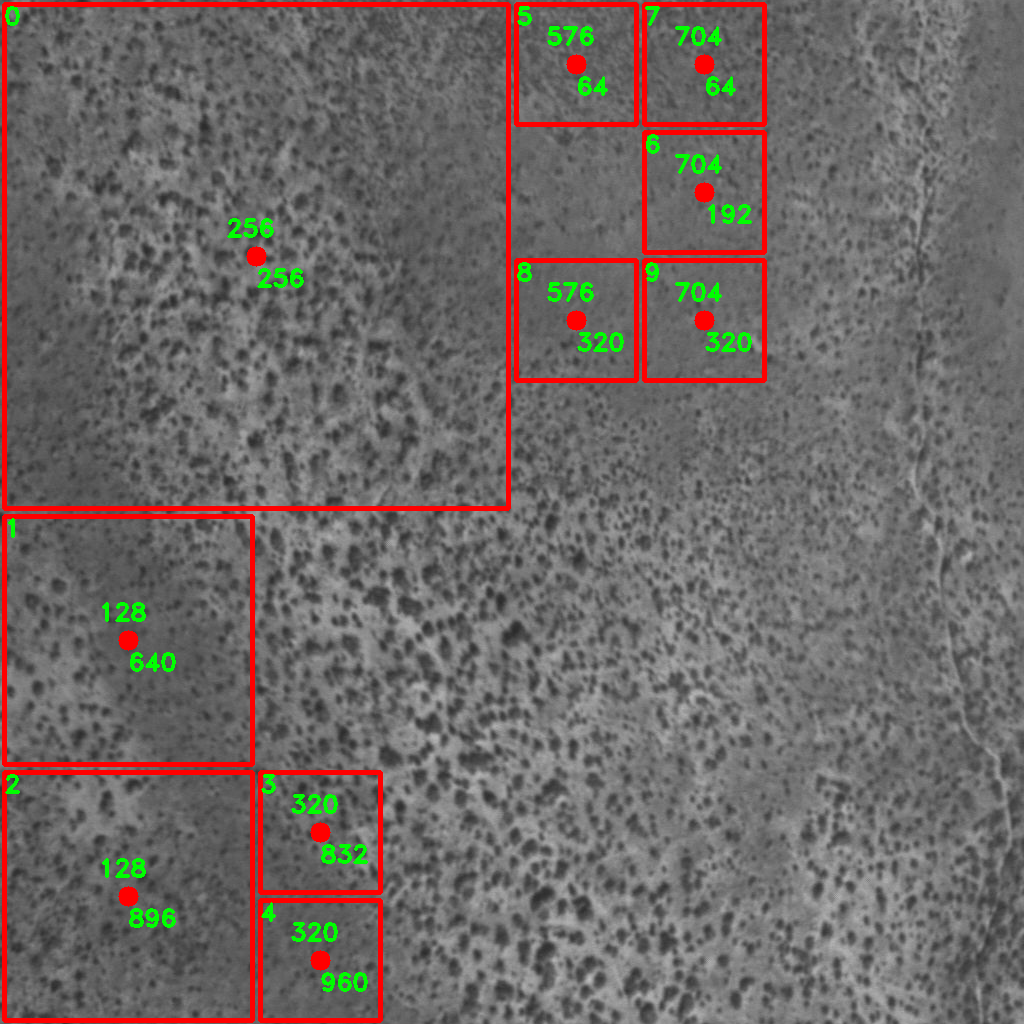}
  \caption{Example quadtree using image from truck testing; the image is of the parking lot concrete from a height of about 4 meters. Note that, in contrast to the previous quadtree result, this quadtree result rejects certain portions of the image due to high standard deviation regardless of the recursive depth.}
  \label{fig:quadtree_airplane}
\end{figure}
\begin{figure}[H]
  \centering
\subfigure[ROI 0] {\label{fig:a1}\includegraphics[width=0.45\textwidth]{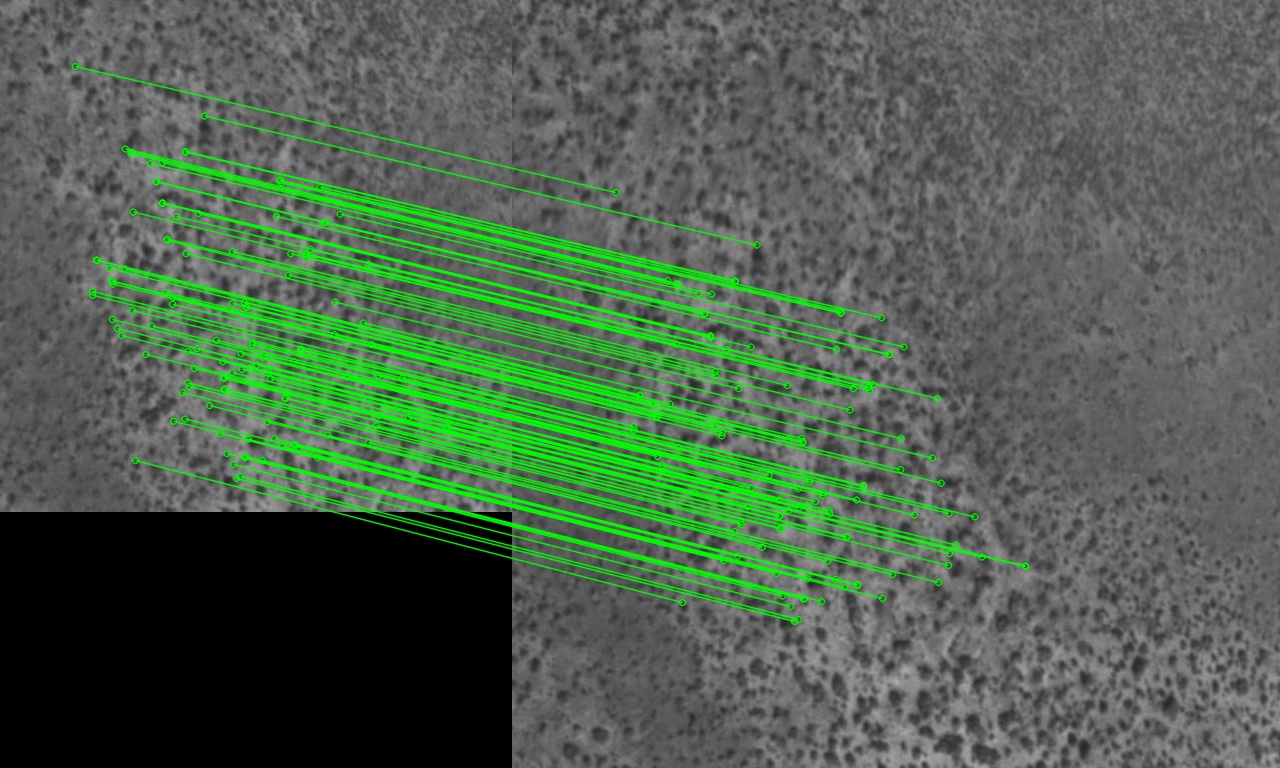}}
\subfigure[ROI 3] {\label{fig:b1}\includegraphics[width=0.45\textwidth]{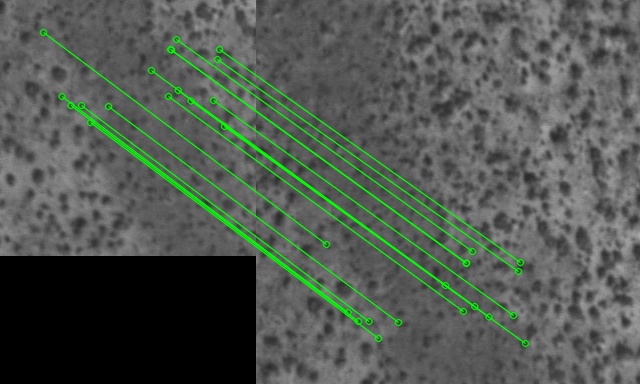}}
\caption{Original keypoints for two different regions prior to outlier rejection through RANSAC and the later nav-based outlier rejection method.}
\label{fig:keypoints_preransac}
\end{figure}

\begin{figure}[H]
  \centering
  \includegraphics[width=\textwidth]{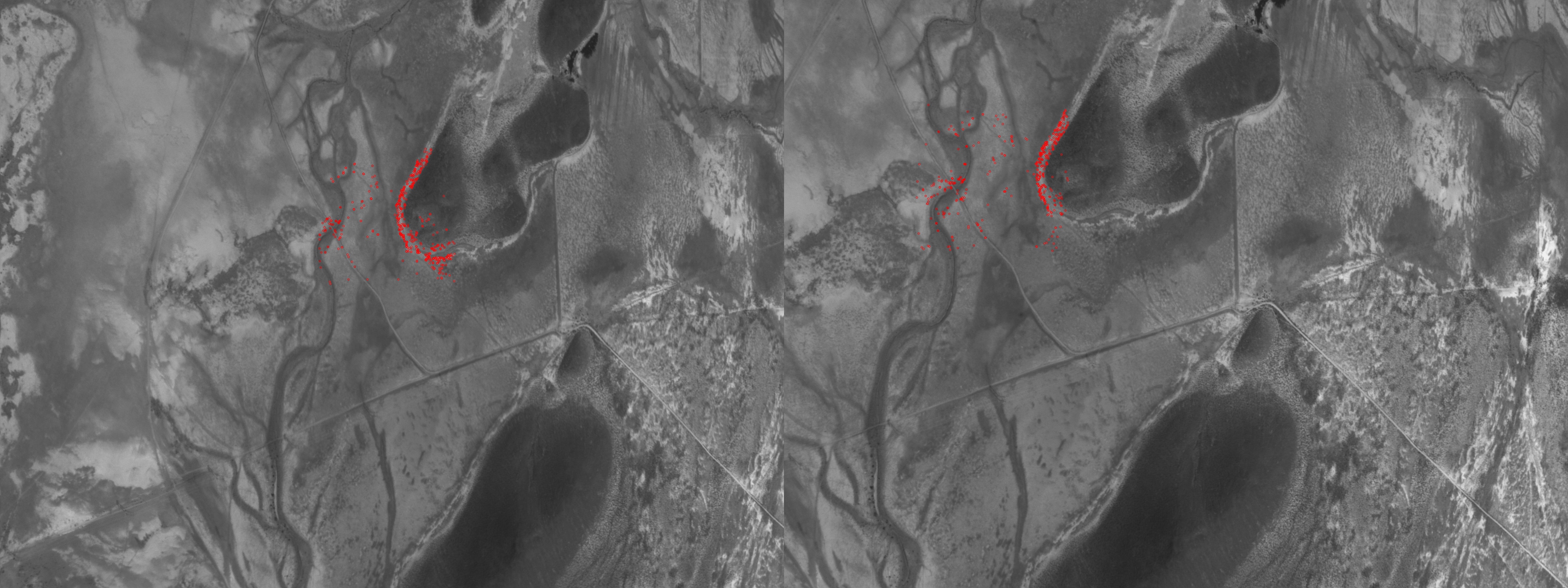}
  \caption{Keypoints for a given ROI in green for first HDA image (left) and second image (right). These keypoints are those keypoints that have survived both the initial RANSAC rejection as well as the outlier rejection based on navigation data. The slope calculated here and for another ROI in the image (ROI 0) both fit the requirements for a landing zone.}
  \label{fig:success_airplane}
\end{figure}

\begin{figure}[H]
  \centering
  \includegraphics[width=0.9\textwidth]{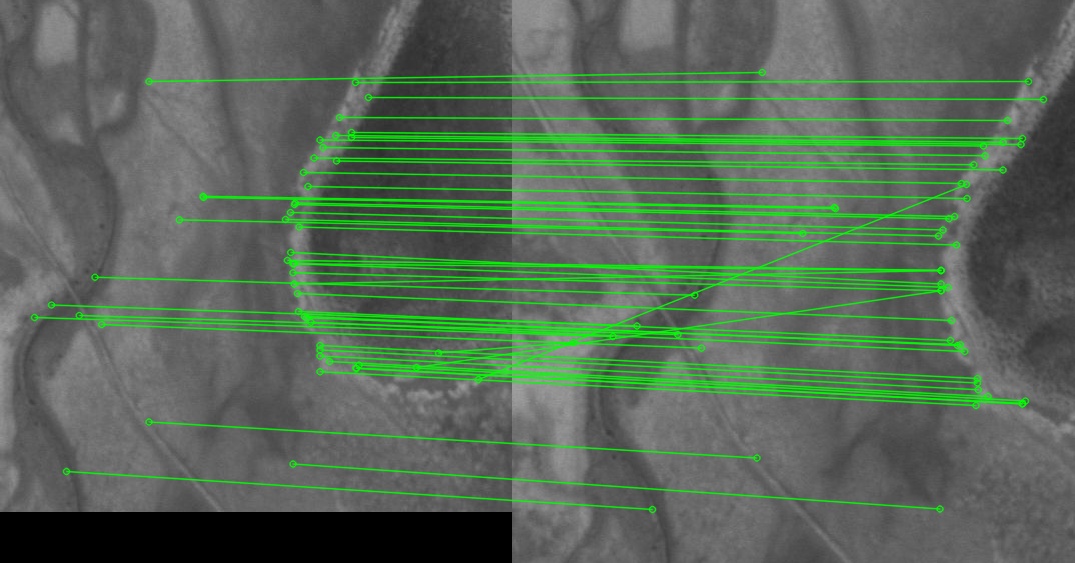}
  \caption{Close up of the two ROIs shown in previous figure. These points are shows after initial RANSAC filtering but prior to nav-based outlier rejection.}
  \label{fig:success_airplane_roi}
\end{figure}

Figure~\ref{fig:success_airplane} shows the result of a flight over a patch of land where the requirements for a landing site have been met---actually the image contains a second ROI that is a possible landing site. The two landing sites have slopes of 4.67 and 5.63 degrees, respectively. You can see the sets of points comprising the detected features in Figure~\ref{fig:success_airplane_roi}.

\begin{figure}
\centering
\subfigure[Ridge without keypoints] {\label{fig:a}\includegraphics[width=0.45\textwidth]{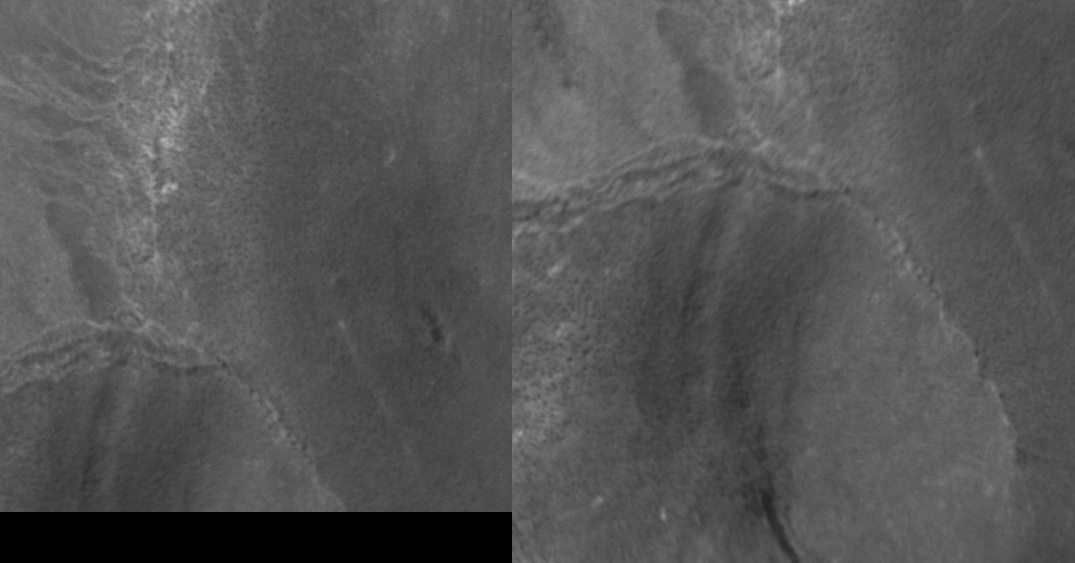}}
\subfigure[Ridge with keypoints] {\label{fig:b}\includegraphics[width=0.45\textwidth]{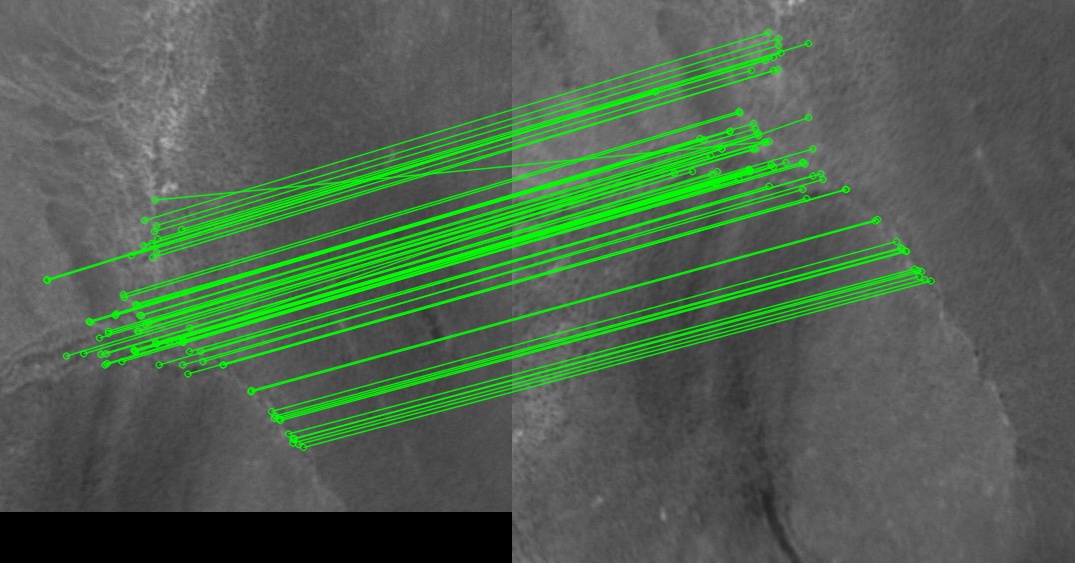}}
\caption{ROIs containing a ridge for point cloud reconstruction.}
\label{fig:roi_ridge}
\end{figure}

\begin{figure}[H]
    \includegraphics[width=\textwidth]{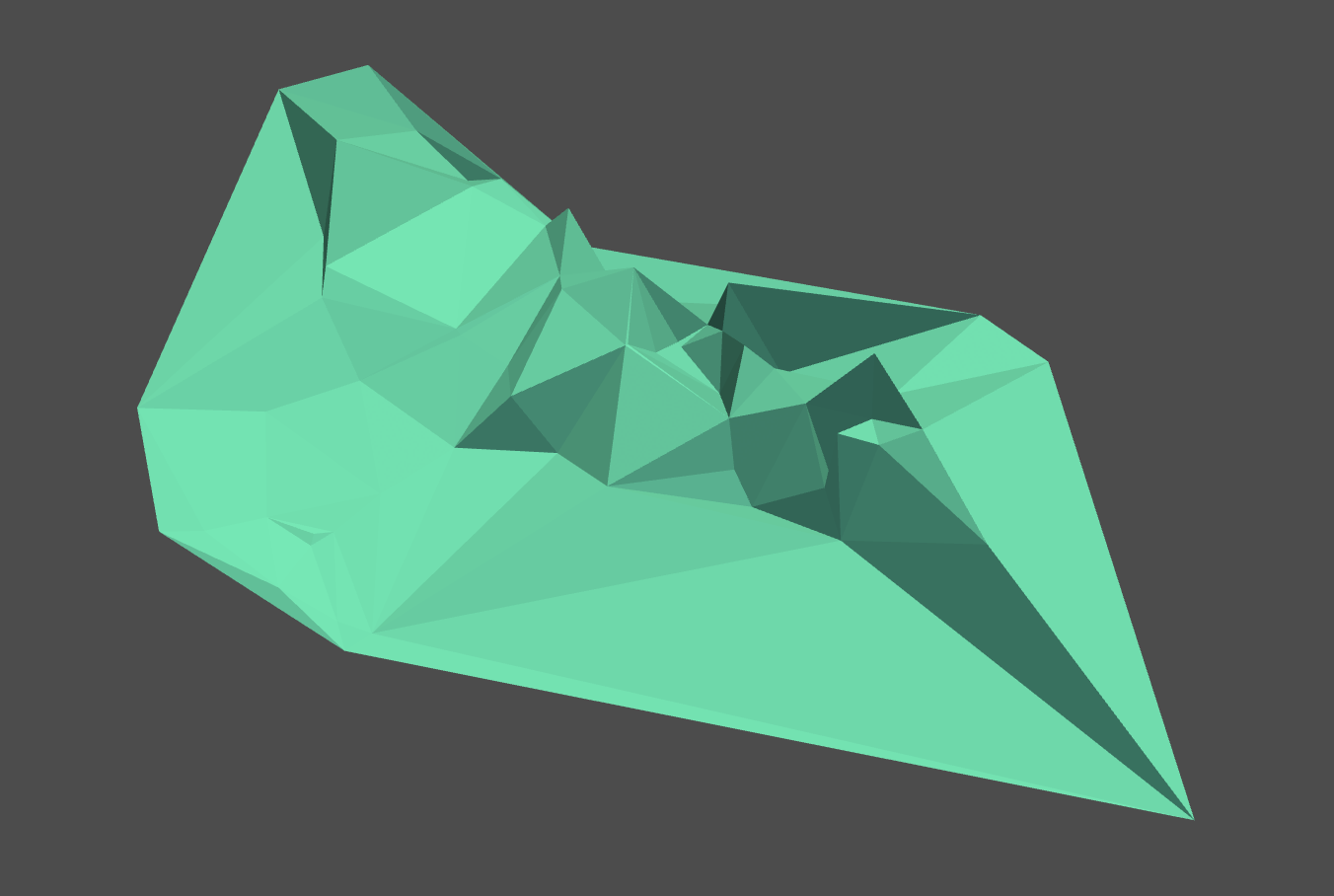}
\caption{Example of point cloud reconstruction from SfM; in a flight over Las Cruces, New Mexico, a ridge is visible in a chosen region, and the resulting point cloud faithfully recreates the general shape of the ridge. Note that this would be a poor landing site, and thankfully the algorithm rejects this ROI. }
\label{fig:ridge_keypoints}
\end{figure}

Even when a point cloud fails to lie within NOVA-C's landing requirements, it can be useful to verify point cloud reconstruction: Figure~\ref{fig:ridge_keypoints} is one such example of an ROI being rejected due to high slopes and Figure~\ref{fig:roi_ridge} shows the regions for which the point cloud was generated.
%
\subsection{Processor in the Loop Campaign}

Historical timing requirements for HDA have often ranged from 1 to 2 minutes\cite{alhatHDA,villalpando2010investigation}.  However, for the NOVA-C lander, a 15 second timing window is required to so guidance and control can maintain sufficient capability to safely land at the designated landing site.  The NOVA-C houses the HDA algorithms on the vision processing unit (VPU) which contains AiTech's SPO-S, a single-board space-rated processor.  Table~\ref{tab:spo} lists a few of the key specifications of the SPO-S computational platform.

\begin{table}[htbp]
	\fontsize{10}{10}\selectfont
    \caption{A Caption Goes Here}
   \label{tab:spo}
        \centering 
   \begin{tabular}{c | r } 
      \hline 
      Item    & Specification  \\
      \hline 
      Processor Speed      & 1 GHz  \\
      RAM          & 1 GB      \\
      OS       & Buildroot Linux (2017)   \\
      \hline
   \end{tabular}
\end{table}

Clearly, this platform is not optimized for the intense needs of vision processing algorithms.  Therefore, regular test cases  are conducted on  engineering units of the VPU.  In addition, profiling data during field testing is continuously collected and analyzed.  To ensure timing constraints are full realized in flight, the HDA flight software contains constant timing checks with kick outs if the algorithm is running long.  Since each candidate landing site is processed by HDA sequentially (see Figure \ref{fig:hda:flowchart}), the affect of a timing kick out is that not all sites would be evaluated.  Therefore, sites closest to the initial landing site are prioritized first.

\subsubsection{Profiling with Synthetic Imagery}

Using synthetic imagery, test cases based on verified SIL cases are ran on the processor with minimal intervention or deviation from the flight configuration.  However, ancillary applications such as navigation are bypassed with a flat file to ensure known and controlled inputs to the HDA application.  This allows for simultaneous flight processor profiling as well as implementation validation.  Profiling with the SIL test described above yields a consistent performance of 2.5 - 3.0 seconds for quadtree, and 6-7 seconds for SfM.

\subsubsection{Profiling of Field Testing}
In the philosophical vein of "test early, test often", profiling HDA during HDA truck testing has been done on all cycles of development.  Truck testing also allows for multiple opportunities to profile the HDA system over various lighting conditions and terrain imagery. Early beta version of truck testing utilized a specialized camera with an image size of 2560$\times$2160 allowed a square image size of 2048$\times$2048, and had some exposed some timing concerns, shown in Table~\ref{tab:pil_spu}. When the SfM algorithm finally ended, both runs lasted around the entire allotted time for the entirety of HDA---if it finished at all which many did not (denoted with an asterisk). Note that this PIL run took place after several attempts at solving this same problem on the target platform and time-saving changes recovered a substantial amount of valuable time. The original several attempts took an average of 50-70 seconds to complete!
\begin{table}[ht]
\caption{PIL Timing Results for USB camera (* denotes Failure)} 
\centering 
\begin{tabular}{c c c c} 
\hline\hline 
Image & Image & Quadtree (s) & SfM (s) \\ [0.5ex] 
\hline 
uvcap\_12279 & uvcap\_12280 & 8.15 & 14.19 \\ 
uvcap\_12279 & uvcap\_12298 & ---- & 2.68* \\ 
uvcap\_12279 & uvcap\_12379 & ---- & 3.74* \\
uvcap\_12279 & uvcap\_12380 & ---- & 2.68* \\
uvcap\_12279 & uvcap\_12384 & ---- & 1.34* \\
uvcap\_12279 & uvcap\_12388 & ---- & 16.08 \\ [1ex] 
\hline 
\end{tabular}
\label{tab:pil_spu} 
\end{table}

After some changes to eliminate unnecessary pieces of the algorithm, the timing changed considerably, as seen in Table~\ref{tab:pil_synthetic}. Note the dramatic drop in processing times, placing us squarely within the timing constraint of 15 seconds even with the time just over ten seconds for SfM, which is more than double the other SfM runs.

\begin{table}[ht]
\caption{PIL Timing Results for synthetic images (* denotes Failure)} 
\centering 
\begin{tabular}{c c c c} 
\hline\hline 
Image 1 & Image 2 & Quadtree (s) & SfM (s) \\ [0.5ex] 
\hline 
00027 & 00032 & 3.44 & 4.90 \\ 
00044 & 00050 & 3.39 & 5.17 \\ 
----- & ----- & 3.34 & 4.73* \\
00071 & 00078 & 3.73 & 4.73 \\
----- & ----- & 3.30 & 10.55* \\
00099 & 00104 & 2.29 & 3.75 \\
00106 & 00116 & 2.78 & 4.17 \\
00117 & 00128 & 2.65 & 3.92 \\ 
00130 & 00140 & 2.63 & 4.44 \\ 
----- & ----- & 2.54 & 3.21* \\ 
[1ex] 
\hline 
\end{tabular}
\label{tab:pil_synthetic} 
\end{table}

After some significant changes and elimination of several unnecessary processes, the following timing constraints were met with the Table~\ref{tab:pil_usb}. Note, however, that these are not completely comparable to timing values in Table~\ref{tab:pil_spu}; the image size used in the testing camera is 2592$\times$1944, meaning that the square image size must be 1024$\times$1024, which is one-quarter the size of the original square image. Nevertheless, even after applying a factor of four to the results in Table~\ref{tab:pil_usb} the updates show a significant and critical reduction in runtimes.
\begin{table}[ht]
\caption{PIL Timing Results for USB camera} 
\centering 
\begin{tabular}{c c c c c } 
\hline\hline 
Image 1 & Image 2 & Run & Quadtree (s) & SfM (s) \\ [0.5ex] 
\hline 
19069 & 19072 & 1 & 0.77 & 1.10 \\ 
19073 & 19075 & 1 & 0.47 & 0.98 \\ 
19077 & 19079 & 1 & 0.46 & 0.97 \\
3651 & 3654 & 2 & 0.87 & 0.49 \\
3656 & 3659 & 2 & 0.49 & 0.85 \\
3661 & 3664 & 2 & 0.49 & 0.82 \\ [1ex] 
\hline 
\end{tabular}
\label{tab:pil_usb} 
\end{table}
%

\subsection{Helicopter Testing Campaign} 
A final hardware in the loop test campaign involving mounting the navpod to a helicopter and flying trajectories at NASA's Kennedy Space Center (KSC) is scheduled for 2022. KSC offers a unique testing with a lunar like terrain and boulder field.  Satisfactory completion of the SIL, PIL, aircraft, and truck testing campaigns is required before advancing to helicopter testing.  Range safety allows for the flight laser range finder unit, so the Navpod config will be as close to flight like as possible.  The glaring exception is that star tracker measurements will not be available, so attitude determination will depend on bearing measurements from surveyed features before take off and gyro propagation during flight.

\section{Conclusion}
We have presented an overview of the HDA design and algorithm for the NOVA-C lander.  We also reviewed an ongoing testing effort that includes SIL, PIL, truck testing, aircraft testing, and future helicopter testing.  Sample results from each testing campaign along with complications were described.  While much work still lies ahead for the team, the NOVA-C team has developed a workable HDA solution that will be verified by mulitple robust and thorough testing campaigns.

\bibliographystyle{AAS_publication}   
\bibliography{hda}   

\begin{thebibliography}{10}

\bibitem{baker2015}
M.~K. Barker, E.~Mazarico, G.~A. Neumann, M.~T. Zuber, J.~Haruyama, and D.~E.
  Smith, ``A new lunar digital elevation model from the Lunar Orbiter Laser
  Altimeter and SELENE Terrain Camera,''  {\em Icarus}, Vol.~273, 2015,
  pp.~346--355, http://dx.doi.org/10.1016/j.icarus.2015.07.039.

\bibitem{esa_landing}
D.~Oddenino, D.~Neveu, J.-F. Hamel, C.~Guerrerio, M.~Mammarella, T.~Voirin,
  A.~M. Barrio, and D.~Beaudette, ``{Scalable GNC Architecture for Moon and
  Mars Landing Missions},''  {\em GNC 2014: 9th International ESA Conference on
  Guidance, Navigation, and Control Systems}, Porto, Portugal, 2014.

\bibitem{epp2007autonomous}
C.~D. Epp and T.~B. Smith, ``{Autonomous precision landing and hazard detection
  and avoidance technology (ALHAT)},''  {\em 2007 IEEE Aerospace Conference},
  IEEE, 2007, pp.~1--7.

\bibitem{crane2014vision}
E.~S. Crane, {\em {Vision-Based Hazard Estimation during Autonomous Lunar
  Landing}}.
\newblock PhD thesis, Stanford University, 2014.

\bibitem{villalpando2010investigation}
C.~Y. Villalpando, A.~E. Johnson, R.~Some, J.~Oberlin, and S.~Goldberg,
  ``{Investigation of the tilera processor for real time hazard detection and
  avoidance on the altair lunar lander},''  {\em 2010 IEEE Aerospace
  Conference}, IEEE, 2010, pp.~1--9.

\bibitem{aaron2022camera}
S.~B. Aaron, Y.~Cheng, N.~Trawny, S.~Mohan, J.~Montgomery, H.~Ansari, K.~Smith,
  A.~E. Johnson, J.~Goguen, and J.~Zheng, ``Camera Simulation For Perseverance
  Rover's Lander Vision System,''  {\em AIAA SCITECH 2022 Forum}, 2022,
  p.~0746.

\bibitem{cheng2021making}
Y.~Cheng, A.~Ansar, and A.~Johnson, ``Making an Onboard Reference Map From
  MRO/CTX Imagery for Mars 2020 Lander Vision System,''  {\em Earth and Space
  Science}, Vol.~8, No.~8, 2021, p.~e2020EA001560.

\bibitem{posada2020a}
D.~Posada, ``An Open Source, Autonomous, Vision-Based Algorithm for Hazard
  Detection and Avoidance for Celestial Body Landing,''  Master's thesis,
  Embry–Riddle Aeronautical University, 2020.
\newblock Master's Thesis. 534.

\bibitem{karami2017image}
E.~Karami, S.~Prasad, and M.~Shehata, ``Image matching using SIFT, SURF, BRIEF
  and ORB: performance comparison for distorted images,''  {\em arXiv preprint
  arXiv:1710.02726}, 2017.

\bibitem{Hartley2004}
R.~I. Hartley and A.~Zisserman, {\em Multiple View Geometry in Computer
  Vision}.
\newblock Cambridge University Press, ISBN: 0521540518, second~ed., 2004.

\bibitem{epipolar}
G.~Xu and Z.~Zhang, {\em Epipolar Geometry in Stereo, Motion, and Object
  Recognition: A Unified Approach}.
\newblock USA: Kluwer Academic Publishers, 1996.

\bibitem{LI201730}
B.~Li, Z.~Ling, J.~Zhang, and J.~Chen, ``Rock size-frequency distributions
  analysis at lunar landing sites based on remote sensing and in-situ
  imagery,''  {\em Planetary and Space Science}, Vol.~146, 2017, pp.~30--39,
  https://doi.org/10.1016/j.pss.2017.08.008.

\bibitem{alhatHDA}
C.~Epp, ``{Autonomous Landing and Hazard Avoidance Technology (ALHAT)},''  {\em
  STAIF 2008: Space Technology and Applications International Forum},
  Albuquerque, NM, February 10-14, 2008.

\end{thebibliography}

\end{document}